\documentclass[runningheads]{llncs}

\usepackage{eccv}

\usepackage{eccvabbrv}

\usepackage{graphicx}
\usepackage{booktabs}

\usepackage[accsupp]{axessibility}  

\usepackage{times}
\usepackage{epsfig}
\usepackage{graphicx}
\usepackage{amsmath}
\usepackage{amssymb}

\usepackage{wrapfig}
\usepackage{blindtext}
\usepackage{multirow}

\usepackage{booktabs}
\usepackage{tabularx}
\usepackage{xcolor,colortbl}
\usepackage{tabularray}
\usepackage{subcaption}
\usepackage{float}
\usepackage{stfloats}

\usepackage[pagebackref,breaklinks,colorlinks,citecolor=eccvblue]{hyperref}

\usepackage{orcidlink}

\begin{document}

\title{Feature Diversification and Adaptation for Federated Domain Generalization} 


\author{Seunghan Yang\orcidlink{0000-0002-0411-8407} \and
Seokeon Choi\orcidlink{0000-0002-1695-5894} \and
Hyunsin Park\orcidlink{0000-0003-3556-5792} \and
Sungha Choi\orcidlink{0000-0003-2313-9243} \and\\
Simyung Chang\orcidlink{0000-0001-7750-191X} \and
Sungrack Yun\orcidlink{0000-0003-2462-3854}}

\authorrunning{S. Yang et al.}

\institute{Qualcomm AI Research$^{\dagger}$ \\
\email{\{seunghan,seokchoi,hyunsinp,sunghac,simychan,sungrack\}\\@qti.qualcomm.com}}

\maketitle

\begin{abstract}
  Federated learning, a distributed learning paradigm, utilizes multiple clients to build a robust global model. In real-world applications, local clients often operate within their limited domains, leading to a `domain shift' across clients. Privacy concerns limit each client's learning to its own domain data, which increase the risk of overfitting. Moreover, the process of aggregating models trained on own limited domain can be potentially lead to a significant degradation in the global model performance. To deal with these challenges, we introduce the concept of federated feature diversification. Each client diversifies the own limited domain data by leveraging global feature statistics, {\it i.e.}, the aggregated average statistics over all participating clients, shared through the global model's parameters. This data diversification helps local models to learn client-invariant representations while preserving privacy.
  Our resultant global model shows robust performance on unseen test domain data. To enhance performance further, we develop an instance-adaptive inference approach tailored for test domain data. Our proposed instance feature adapter dynamically adjusts feature statistics to align with the test input, thereby reducing the domain gap between the test and training domains. We show that our method achieves state-of-the-art performance on several domain generalization benchmarks within a federated learning setting.

\end{abstract}

\section{Introduction}
\label{sec:intro}
{\let\thefootnote\relax\footnotetext{{
$\dagger$ Qualcomm AI Research is an initiative of Qualcomm Technologies, Inc.}}}
Training a deep learning model with direct access to distributed data presents significant privacy concerns. Consequently, federated learning (FL)~\cite{SiloBN, hosseini21a, SCAFFOLD, FedProx, FedBN, FedAvg, FedNova, yang2022robust} has surfaced as a promising solution for avoiding direct data access. Early FL research like FedAvg~\cite{FedAvg} and FedProx~\cite{FedProx} tackled privacy by exchanging locally trained model parameters, rather than the actual data of local clients. These parameters are then aggregated on a central server to generate a global model, which is returned to the clients to continue training until stability is reached.

\begin{figure}[t]
\begin{center}
\epsfig{figure=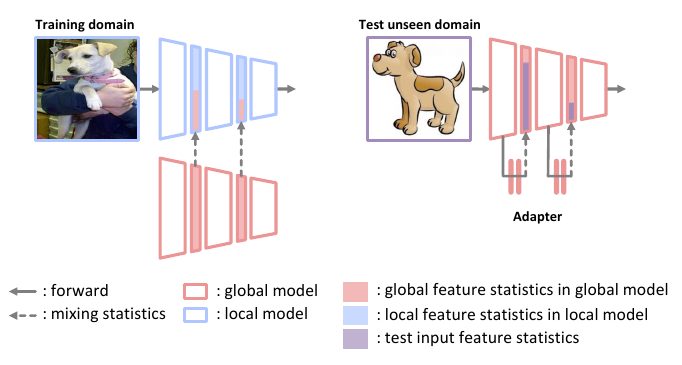}
\end{center}
   \caption{
   {\bf Our Concept Diagram.} Federated feature diversification: during training, local data is diversified using global feature statistics—averaged statistics across all client data—provided by the global model. This approach aids in learning representations that are invariant to individual client characteristics.
   Instance feature adaptation: in the test phase, the adapter employs input feature statistics to facilitate instance-adaptive inference tailored for new and unseen test data.}
\label{fig1}
\end{figure}

Contrary to common assumptions in previous studies that local data across clients originates from a uniform source domain, in practice, each client's local data often derives from its own distinct source domain. This leads to a domain disparity among clients. For example, in autonomous driving scenarios, disparity arises as each vehicle captures street views and infrastructure with variations attributable to its specific camera sensors, geographic location, and other factors. Such variation in data distribution between clients results in what is known as a `domain shift,' which poses a prevalent challenge in both the training and testing phases of model development.

This paper addresses these problems by exploring federated domain generalization (federated DG), which aims to improve a federated model's learning process in handling various distributed source domains while maintaining reliable generalization to unseen domains. While past studies like SiloBN~\cite{SiloBN} and FedBN~\cite{FedBN} have aimed to manage the domain shift within distributed source domains, they do not tackle the domain shift that occurs between training and test distributions.
The attempts to enhance generalizability to unseen domains, such as FedDG~\cite{FedDG}, COPA~\cite{COPA}, FedIns~\cite{FedINS}, CCST~\cite{CCST}, CSAC~\cite{CSAC}, GA~\cite{GA}, and TsmoBN~\cite{TsmoBN}, have been made, but these methods have their limitations. These include potential privacy leaks through shared client information~\cite{FedDG, COPA, FedINS, CCST}, limited performance by prioritizing aggregation over local training~\cite{CSAC, GA} and the necessity of target data for adaptation to the test domain~\cite{TsmoBN}. Unlike previous works, we resolve federated DG without causing additional privacy leaks and without utilizing target data in learning.

In this study, presented in Figure~\ref{fig1}, we propose a novel framework, named {\it FedFD-A}, which comprises two approaches: federated feature diversification and instance feature adaptation.
Our first approach, federated feature diversification, obtains global feature statistics with the average statistics over all client data and diversifies the limited local data using the obtained global feature statistics.
Specifically, we access global feature statistics through batch normalization layers in the server model and apply a combination of local and global feature statistics to normalize local data, thereby achieving augmented features. Training with these data mitigates the overfitting of each local model to its specific domain by learning client-invariant representations.
Previous approaches, such as FedDG~\cite{FedDG}, directly access other clients' domains for data diversification, while methods like MixStyle~\cite{Mixstyle} are limited to diversifying data within the same batch, where samples are collected from the same domain on the client. In contrast, our method leverages information shared with the server model. Our method maintains privacy while still effectively leveraging information across various domains.

Our second approach, instance feature adaptation, aims to enhance generalization to unseen targets. In the federated learning stage, the adapter is trained to leverage input feature statistics for mitigating domain shift between the input and the training distribution. We achieve this by normalizing the input using a combination of input and global statistics, interpolated through the adapter. At the testing stage, the adapter employs feature statistics from the test data to reduce the domain shift without needing further adjustments, unlike TsmoBN~\cite{TsmoBN}, which requires additional adaptation steps during inference.

We evaluate our FedFD-A components through extensive experiments on several DG benchmarks in the image classification task, including PACS~\cite{PACS}, VLCS~\cite{VLCS}, OfficeHome~\cite{OfficeHome}, and DomainNet~\cite{DomainNet}.
Our framework consistently enhances the performance of the federated model across multiple domain shift scenarios.

\section{Related Works}

\subsection{Federated Learning}

Federated learning (FL) has gained substantial traction due to its ability to train a global model using decentralized datasets, all while maintaining user privacy. Most recent FL methodologies have focused on addressing the issue of non-iid data spread across clients, specifically those with heterogeneous label distributions~\cite{SCAFFOLD, FedProx, FedNova}. One notable method, FedProx~\cite{FedProx}, reduces the disparity between local and global models by incorporating a proximal term into local loss functions to regularize these local models.
However, the issue of non-iid feature distribution has not garnered as much attention in previous studies. SiloBN~\cite{SiloBN} and FedBN~\cite{FedBN} strive to manage the domain shift by preserving batch normalization (BN) statistics at a local level rather than aggregating them server-side. TsmoBN~\cite{TsmoBN} aims to boost performance by updating test BN to adapt the global model towards target clients, provided they possess a substantial amount of data in the target domain.
Distinct from previous studies, our research leverages the entirety of client feature statistics, allowing the effective training of local models to learn client-invariant representations. Furthermore, we introduce an instance adaptation strategy, which enables the global model to generalize directly to unseen domains using test input data alone.

\subsection{Domain Generalization}
Domain generalization aims to train a model on source domains in a manner that facilitates robust performance on unseen target data. Single-source approaches~\cite{JiGen, RSC, SelfReg, SFA, L2D, RandConv, Mixstyle, ProRandConv} can be applied within a FL context while preserving privacy. However, these methods can underutilize the rich server information inherent in FL, thereby offering only marginal performance improvements.
On the other hand, multi-source methods~\cite{CompoundDG, chen2021style, EpiFCR, lv2022causality, BIN, pandey2021Generalization, DSON, FourierDG, MBMSML} that minimize domain discrepancies across sources necessitate raw data sharing, prompting privacy concerns within FL. Some recent studies, such as FedDG~\cite{FedDG}, COPA~\cite{COPA}, and CCST~\cite{CCST}, have utilized multi-source techniques in distributed settings without sharing raw data; nonetheless, they still share potentially sensitive style distributions or classifiers.
The CSAC~\cite{CSAC} approach mitigates the domain shift via an aggregation method but requires pre-training and results in a limited performance increase. Similarly, GA~\cite{GA} aggregates local models taking into account client divergences, but the improvements achieved are generally marginal. FedIns~\cite{FedINS} provides dynamic global model adjustments per testing instance, but necessitates sharing elements of local models amongst clients, potentially compromising privacy.
Contrarily, our study introduces a feature diversification methodology that addresses the domain shift without causing additional privacy issues. Moreover, our instance-adaptive feature adaptation approach directly generalizes the global model to accommodate testing instances without needing to share local models.

\section{Proposed Methods}

\subsection{Preliminaries}
{\noindent\bf Notation and Problem Formulation:}
Let $\mathcal{X}$ and $\mathcal{Y}$ denote the input and label spaces, respectively.
The $k$-th client has a single-domain data $D_{k}=\{(x_{i, k}, y_{i, k})\}^{n_{k}}_{i=1}$, where $n_k$ is the number of samples. The set of distributed source domain data from $K$ clients is represented as $\{D_{1}, ..., D_{K}\}$.
In federated domain generalization (federated DG), there exists a domain shift across clients, where each client data $D_{k}$ sampled from a domain-specific distribution $(\mathcal{X}_{k}, \mathcal{Y})$ that differs from other clients.
The target test domain data from an unseen environment is represented as $D_{t}$, with a distribution $(\mathcal{X}_{t}, \mathcal{Y})$ that is shifted from the training data.
The feature extractor, parameterized by $\theta$, is represented as $F_{\theta}$, and the classifier, parameterized by $\phi$, is represented as $C_{\phi}$.
Federated DG aims to learn a generalized global model $C_{\phi_{G}} \circ F_{\theta_{G}} : \mathcal{X} \rightarrow \mathcal{Y}$ by aggregating $K$ distributed clients' models $\{F_{\theta_{k}}, C_{\phi_{k}}\}_{k=1}^{K}$ trained on source data $\{D_{k}\}_{k=1}^{K}$, such that the global model generalizes to unseen target domain $D_{t}$.
Note that we exploit FedAvg~\cite{FedAvg} as an aggregation method, where the global model parameters are weighted averaged with the local model parameters based on the dataset size $\theta_{G} = \sum_{k=1}^{K} \frac{n_{k}}{n}\theta_{k}$ and $\phi_{G} = \sum_{k=1}^{K} \frac{n_{k}}{n}\phi_{k}$. Here, $n$ indicates the total number of data across clients.

{\noindent\bf Batch Normalization:}
In the context of this study, batch normalization (BN) layers operate as $\gamma^{l}_{k} \frac{a^{l}_{k}-\mu^{l}_{k}}{\sigma^{l}_{k}}+\beta^{l}_{k}$. Here, $\mu^{l}_{k}$ and $\sigma^{l}_{k}$ represent the mean and standard deviation statistics for the $l$-th BN layer in the $k$-th client.
These statistics reflect the overall feature statistics from data in the $k$-th local client by calculating a running mean and standard deviation.
Additionally, $a^{l}_{k}$ refers to an input tensor, while $\gamma^{l}_{k}$ and $\beta^{l}_{k}$ signify learnable scaling and shifting parameters.
In this work, we employ the terms `local statistics' to refer to $\mu_{k}$ and $\sigma_{k}$, while `global statistics' represents $\mu_{G}$ and $\sigma_{G}$, which are computed by aggregating the local statistics.

\begin{figure*}[ht]
\centering
\epsfig{figure=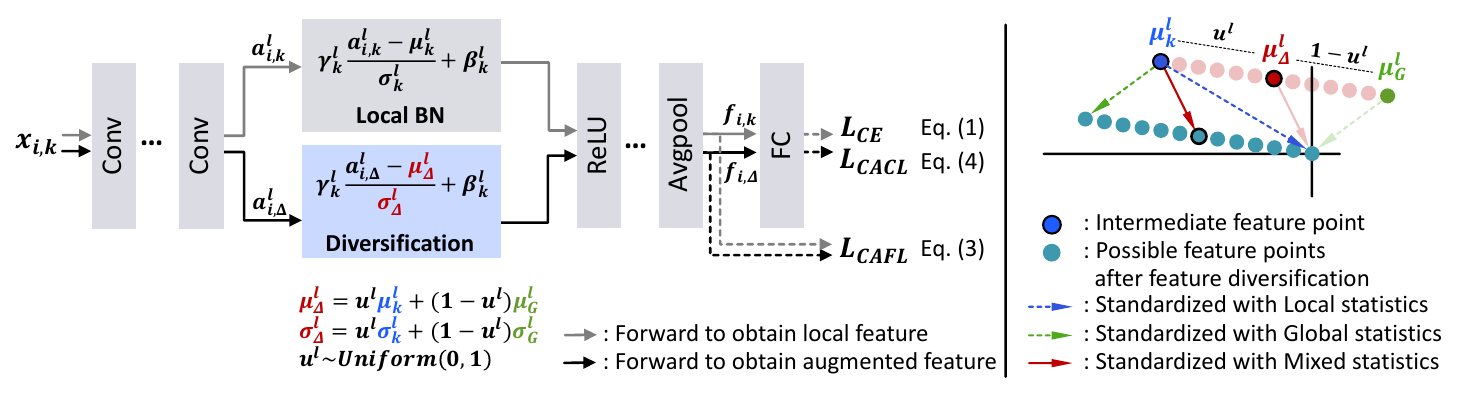,width=\linewidth}
	\caption{
 {\bf Federated Feature Diversification} begins with obtaining the local feature $f_{i,k}$ using batch normalization (BN) at the individual client level.
 Simultaneously, the augmented feature $f_{i,\Delta}$ is derived through mixed statistics, implemented by interpolating local and global statistics randomly.
 The intermediate feature ${a}^{l}_{i,\Delta}$ is then standardized with these statistics, leading to a range of features (as displayed on the right figure). This procedure is consistently applied across all BN layers to produce a greater variety of features. These diversified features are then employed to learn client-invariant representations, as specified by Eq.~\ref{eq44} and~\ref{eq5}.
 }
\label{fig2}
\end{figure*}

\subsection{Federated Feature Diversification}
In local training, the $k$-th local model is trained with the cross-entropy loss function $\text{CE}(\cdot)$ on its dataset as follows:
\begin{equation}
    \mathcal{L}_{CE} = \frac{1}{n_{k}} \sum_{i=1}^{n_{k}} \text{CE}(C_{\phi_{k}}(F_{\theta_{k}}(x_{i,k})), y_{i,k}).
\label{ce_eq}
\end{equation}
Before starting local training, each client receives the global model parameters $\{\theta_{G}, \phi_{G}\}$. Using these parameters, the local model $\{\theta_{k}, \phi_{k}\}$ is initialized. Subsequently, $F_{\theta_{k}}$ and $C_{\phi_{k}}$ are trained on its data during local epochs.
Despite initializing the local model with the global model's parameters, which are generally robust, a challenge remains. The problem emerges when the local model relies solely on the cross-entropy loss from a single domain, mostly due to privacy concerns associated with using data from other clients directly.
This approach can lead to overfitting to the data of the single domain, causing a significant divergence in the local client models. Such divergence can pose difficulties in combining the local models to form the generalized global model as noted in literature~\cite{FedBN, SiloBN}.
To mitigate this issue, we propose training the local model with data that is diversified across multiple domains while maintaining privacy of the clients. This varied data is created using our proposed method of data diversification.

{\noindent\bf Diversification Process:}
In federated learning, we accumulate both local and global statistics through BN layers. These global statistics reflect the feature statistics across all clients—a property we utilize for data augmentation during local training, as shown in Fig.~\ref{fig2}.
In conventional local training, the input tensor $a^{l}_{i,k}$ is normalized using batch samples' statistics in the $l$-th BN layer. The local feature $f_{i,k}$ is then consistently calculated based on statistics derived from single domain local data.
Such an approach restricts the local model to learn representations only within a single domain.
We propose the resolution of this limitation by normalizing input tensors with a mix of statistics–exploiting global and local statistics concurrently.
These mixed statistics are calculated as follows:
\begin{equation}
    \mu^{l}_{\Delta} = u^{l}\mu^{l}_{k} + (1 - u^{l})\mu^{l}_{G} \; \text{and} \; \sigma^{l}_{\Delta} = u^{l}\sigma^{l}_{k} + (1 - u^{l})\sigma^{l}_{G}.
\label{eq4}
\end{equation}
We define $u^{l} \in R^{C^{l}}$ as an interpolation weight vector, where each element is independently drawn from uniform distribution $U(0,1)$ with every iteration. Here, $C^{l}$ is the feature dimension in the $l$-th BN layer.
The local statistics can be calculated on a batch or instance basis. However, we opt to use instance-based statistics to generate a more diverse range of domains.

Features normalized by local statistics reflect local characteristics, while features normalized by global statistics incorporate global representations. Random interpolation of these two types of statistics across all BN layers generates a rich diversity of data that fully exploits the features of both local and global domains.
In Fig.~\ref{fig2}, we denote the intermediate feature by ${a}^{l}_{i,\Delta}$ normalized with mixed statistics, and the augmented feature $f_{i, \Delta}$ is obtained with $\{\mu^{l}_{\Delta}, \sigma^{l}_{\Delta}\}_{l=1}^{L}$.
We train the local model using both $f_{i,k}$ and $f_{i, \Delta}$ in a client-agnostic way, which is described in the following section.
In contrast to previous works~\cite{SFA, Mixstyle} that augment features using random noise values or styles of batch samples, our approach employs aggregated feature statistics—safely creating diverse augmentations in line with multi-source domain generalization.
Note that our method uses global statistics in the server, which can reduce privacy leakage.

{\noindent\bf Client-agnostic Learning Objectives (CAL):}
We utilize a client-agnostic feature loss (CAFL) function as follows:
\begin{equation}
    \mathcal{L}_{CAFL} = \frac{1}{n_{k}} \sum_{i=1}^{n_{k}}\left \| f_{i, k} - f_{i, \Delta} \right \|_{2}^{2}.
\label{eq44}
\end{equation}
This loss function forces the local model to learn client-invariant representations, which is achieved by minimizing the variation between original and diversified features.
Further, we train the local classifier to categorize the diversified features, thus encouraging the classifier to disregard client-specific information.
This is achieved through our client-agnostic classification loss (CACL) function as follows:
\begin{equation}
    \mathcal{L}_{CACL} = \frac{1}{n_{k}} \sum_{i=1}^{n_{k}}\text{CE}(C_{\phi_{k}}(f_{i, \Delta}), y_{i,k}).
\label{eq5}
\end{equation}
Our client-agnostic learning functions as a regularization method, which restrains local models from diverging significantly from the global model.
Different from the previous work~\cite{FedProx} that directly regularizes local weight parameters in line with the global model, our proposed learning evaluates the significance of weight parameters for client-invariant representations within diverse domain data.
Finally, the loss for local optimization is provided as follows:
\begin{equation}
    \mathcal{L}_{total} = (1 - \lambda_{1}) \cdot \mathcal{L}_{CE} + \lambda_{1} \cdot \mathcal{L}_{CACL} + \lambda_{2} \cdot \mathcal{L}_{CAFL},
\label{eq7}
\end{equation}
where $\lambda_{1}$ and $\lambda_{2}$ are balancing parameters.

\begin{figure}
\centering
	\epsfig{figure=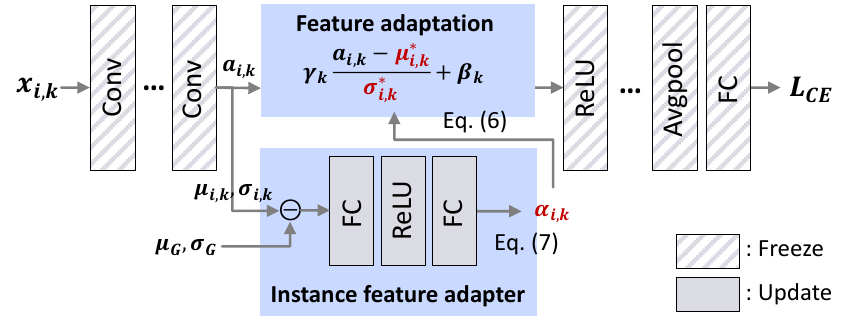,width=0.7\textwidth}
	\caption{
	{\bf Instance Feature Adaptation.} The instance feature adapter takes the difference between instance and global statistics and generates interpolated statistics through the Eq.~\ref{interpolated_bn} and~\ref{eq10}. This procedure is consistently applied across all BN layers to effectively bridge the domain gap.
	}
\label{fig3}
\vspace{-3mm}
\end{figure}

\subsection{Instance Feature Adaptation}
At test time, it can be difficult to generalize to completely unseen domains where the data distribution has shifted from the training set. Addressing this domain discrepancy, we propose utilizing test data statistics, drawing inspiration from domain generalization and test-time adaptation methods.
Previous domain generalization methods attempted to bridge the domain gap using the most relevant BN layer ~\cite{CompoundDG, DualNorm} with the test input among multiple source BN layers. However, these methods are not feasible in a federated setting in which local models can only access the global, not the other clients' BN layers due to privacy constraints.
Moreover, recent test-time adaptation methods ~\cite{AlphaBN, MixNorm, IABN} use interpolated statistics between instance and learned statistics to reflect the test input distribution, yet they require manual or rule-based generation of interpolation parameters. Instead, we propose a learning-based network to dynamically generate instance-wise interpolation parameters for interpolated statistics.

{\noindent\bf Interpolated BN Statistics:}
To simplify the notation, we describe only the process at the $l$-th BN layer and omit the notation related to the sequence of BN layers.
We leverage both instance statistics from the test input $x_{i, t}$ and global statistics as follows:
\begin{equation}
    \mu^{*}_{i, t} = \alpha_{i, t} \mu_{i, t} + (1-\alpha_{i, t}) \mu_{G} \; \text{and} \; \sigma^{*}_{i, t} = \alpha_{i, t} \sigma_{i, t} + (1-\alpha_{i, t}) \sigma_{G},
\label{interpolated_bn}
\end{equation}
where $\mu_{i, t}$ and $\sigma_{i, t}$ are the instance mean and standard deviation of the input tensor.
$\mu_{G}$ and $\sigma_{G}$ reflects the overall feature statistics of local clients involved in training.
Our method leverages $\mu^{*}_{i, t}$ and $\sigma^{*}_{i, t}$ to normalize the test input tensor, with $\alpha_{i, t}$ serving as an interpolation parameter that modulates the impact of instance statistics.
Ideally, the optimal $\alpha_{i, t}$ would be determined upon the availability of the test domain, but this information is typically inaccessible.
Consequently, we propose an instance adaptation that dynamically produces $\alpha_{i, t}$ for each test input in a careful and well-designed manner.

{\noindent\bf Design of Instance Adapter:}
We construct an instance adapter $G_{\varphi_{G}}$, parameterized by $\varphi_{G}$, and integrated into each BN layer in the feature extractor.
The adapter takes the channel-wise distance between instance and global statistics as its input and generates $\alpha_{i, t}$. 
This design is motivated by the out-of-distribution (OOD) detection work ~\cite{dong2022neural}, which leverages the distance between test inputs and learned statistics for OOD sample selection.
Through this design, the adapter generates the appropriate interpolation parameters based on the discrepancy between the test and training distribution of each layer.
Here, we set $\alpha_{i, t}$ as a scalar value to interpolate instance and global statistics with the same weights along the channel axis in the BN layer. It reduces the network size of the adapter and mitigates overfitting to the local data.

{\noindent\bf Training Process:}
In the $k$-th client, we simulate test conditions where instance statistics differ from global statistics. Instead of using test samples, we input the training sample $x_{i, k}$ into the main network, $F_{\theta_{k}}$ and $C_{\phi_{k}}$. As shown in Fig.~\ref{fig3}, BN statistics are replaced with interpolated BN statistics in Eq.~\ref{interpolated_bn}, where $\alpha_{i, k}$ is generated by a local adapter $G_{\varphi_{k}}$ based on the difference between instance and global statistics.
We employ $\mathcal{L}_{CE}$ to train the adapter, enabling it to generate the optimal interpolation parameters in an instance-adaptive fashion.
Although the training data are more similar to global statistics compared to the test data, leveraging their differences teaches the adapter how to balance these statistics.
Here, the main network focuses on generalization through feature diversification, while the adapter further adapts to unseen test domain data via instance feature adaptation.
Thus, we implement alternating training to maintain this separation.

Moreover, to prevent over-fitting of the adapter, we apply the reparameterization trick~\cite{Reparameterization}, generating interpolation parameters by sampling from gaussian distribution reparameterized by the adapter:
\begin{equation}
    \alpha_{i, k} = \mathrm{T}(z\cdot\delta_{i, k} + \epsilon_{i, k}), \; \text{where} \; [\delta_{i, k};\epsilon_{i, k}] =  G_{\varphi_{k}}([\mu_{i, k}-\mu_{G};\sigma_{i, k}-\sigma_{G}]),
\label{eq10}
\end{equation}
with $z$ sampled from $N(0,1)$. $\mathrm{T}(\cdot)$ is a clamp function to ensure $\alpha_{i, k}$ within the range $[0, 1]$. $[\mu_{i, k}-\mu_{G};\sigma_{i, k}-\sigma_{G}]$ represents the concatenation of the difference between instance and global statistics along the channel axis.
During the federated learning stage, the local adapter $G_{\varphi_{k}}$ is trained on each client's data, and these local adapters are subsequently aggregated into the server adapter $G_{\varphi_{G}}$.

{\noindent\bf Test Process:}
We use the server adapter $G_{\varphi_{G}}$ to get $\delta_{i, t}$ and $\epsilon_{i, t}$. In Eq.~\ref{eq10}, we can then calculate $\alpha_{i, t}$ as $\mathrm{T}(\epsilon_{i, t})$ because the mean of $z$ is $0$.
In every BN layer, normalization is carried out using the interpolated statistics in Eq.~\ref{interpolated_bn}, and the procedure requiring only single forward propagation. The computational cost is evaluated in the experimental section. FedFD refers to the method using feature diversification, while FedFD-A denotes the approach that also utilizes the adapter. The pseudo-code of our algorithm can be found in supplementary material.

\section{Experiments}
\subsection{Experimental Setup}
\label{experimental_setup}
{\noindent\bf Datasets and Settings:} 
We evaluate the performance of our FedFD-A on four benchmarks in the domain generalization setting of image classification. We use the PACS~\cite{PACS}, VLCS~\cite{VLCS}, OfficeHome~\cite{OfficeHome}, and DomainNet~\cite{DomainNet} datasets. PACS contains seven categories from Photo (P), Art painting (A), Cartoon (C), and Sketch (S) domains, notable for their substantial domain shifts~\cite{CompoundDG, domain_shift1}. VLCS consists of five categories from VOC2007 (V)~\cite{VOC2007}, LabelMe (L)~\cite{LabelMe}, Caltech-101 (C)~\cite{Caltech}, and SUN (S)~\cite{SUN} domains, primarily experiences domain shifting due to camera type variations.
Comparatively, OfficeHome~\cite{OfficeHome}, which includes 65 categories from Artistic (A), Clipart (C), Product (P), and Real world (R) domains, faces a less drastic issue of the domain shift~\cite{domain_shift2, COPA}.
Consistent with prior federated DG research~\cite{FedDG, COPA, CSAC}, we assume that each client has data from a single domain and that there are a total of four clients in each domain generalization benchmark. The global model is collaboratively learned with three of the clients and is then evaluated on the remaining domain.

For a more comprehensive understanding of our FedFD-A in practical settings, please refer to supplementary material. This includes further experiments on federated DG with the DomainNet~\cite{DomainNet} dataset, exploration of an increased number of clients, and scenarios where non-iid label distribution is also present. These scenarios provide valuable insights into the realistic application of our model.



{\noindent\bf Implementation Details:}
In the federated DG scenario, all clients employ the same architecture and hyper-parameter configurations. The backbone network is a ResNet-18~\cite{ResNet} model pre-trained on ImageNet. In FedFD-A, two fully connected layers are introduced before each BN layer to serve as the instance adapter, where output size is $2$. The network is trained via Stochastic Gradient Descent (SGD) with a fixed learning rate of $0.01$, momentum of $0.5$, and batch size of $64$ over $200$ iterations per round.
We conduct $40$ rounds of training. The values of $\lambda_{1}$ and $\lambda_{2}$ in Eq.~\ref{eq7} are set to $0.1$ and $4.0$, respectively.
Ablation studies of balancing parameters are presented in supplementary material.

{\noindent\bf Evaluation Protocols:}
In our evaluation of the federated DG performance, we adopt the standard protocols specified in DomainBed~\cite{Domainbed} for dataset splits and model selection strategies.
Specifically, we use the training-domain validation set for model selection.
To obtain the best global model in the federated DG setting, different selection strategies are adopted for the client side and the server side.
On the client side, the local model is uploaded to the server when the validation accuracy on its local domain is best within $200$ iterations, validating every $20$ iterations.
On the server side, the best round model is selected when the average validation performance on participating clients is maximized among $40$ rounds, validating every round.
The server can access clients' validation performance using the aggregated server parameters, making it possible to use the average validation performance on participating clients. This strategy of adopting single-source DG validation on local training and multi-source DG validation on the server is both practical and effective.
To ensure a fair comparison, we reproduce competitive methods in our federated DG setting. The results are reported as the average accuracy and standard deviation over four runs with different random seeds.

{\noindent\bf Baseline Frameworks:}
{Our FedFD-A is designed to be compatible with a range of federated learning frameworks. Specifically, we have applied our approach in conjunction with FedBN~\cite{FedBN} and SiloBN~\cite{SiloBN}, both of which aim to mitigate the domain shift by preserving batch normalization (BN) parameters locally. FedBN retains both BN statistics and affine parameters, while SiloBN only preserves BN statistics, thus enabling the BN layer to implicitly learn client-specific information.
Note that parameters are aggregated for the global model with the same process of FedAvg~\cite{FedAvg}.
In Table~\ref{baselines}, we demonstrate that our proposed FedFD-A method significantly improves the performance of baseline methods by approximately 5-7\%. Among the methods, SiloBN is particularly effective in reducing the domain gap and exhibits the highest performance, as reported in~\cite{feddrive}. For the remainder of our experiments, we focus solely on the SiloBN framework.}

\begin{table}[t]
\centering
\caption{Accuracy (\%) on PACS using FedFD-A on the baselines.}
\begin{minipage}{0.2\textwidth}
\centering
\resizebox{0.9\linewidth}{!}{%
\begin{tabular}{c|c}
\toprule
Methods & Acc.  \\ \midrule
FedAvg~\cite{FedAvg} &  77.35  \\
+ FedFD-A &  84.22  \\\bottomrule
\end{tabular}}
\end{minipage}%
\hspace{8mm}
\begin{minipage}{0.2\textwidth}
\centering
\resizebox{0.9\linewidth}{!}{%
\begin{tabular}{c|c}
\toprule
Methods & Acc.  \\ \midrule
FedBN~\cite{FedBN} &  79.57  \\
+ FedFD-A &  85.15  \\\bottomrule
\end{tabular}}
\end{minipage}%
\hspace{8mm}
\begin{minipage}{0.2\textwidth}
\centering
\resizebox{0.9\linewidth}{!}{%
\begin{tabular}{c|c}
\toprule
Methods & Acc.  \\ \midrule
SiloBN~\cite{SiloBN} &  80.79  \\
+ FedFD-A &  85.53  \\\bottomrule
\end{tabular}}
\end{minipage}
\label{baselines}
\end{table}

\subsection{Ablation Studies of Our Components}
{\noindent\bf Variants of Federated Feature Diversification:}
We perform experiments to examine the properties of our feature diversification approach in Table~\ref{ablation1}. SiloBN is used as the baseline, and two client-agnostic learning objectives are applied using diversified data. In this experiment, we excluded our adapter to focus solely on diversification analysis.

When solely global statistics are utilized, set as $u=0.0$ in Eq.~\ref{eq4}, the inputs are normalized using the statistics of the global model.
These global statistics encapsulate the feature statistics across all training clients. By augmenting local data through normalizing with these statistics, the model is primed to learn domain-invariant representations.
Conversely, when utilizing only instance statistics, the style-removed features, proposed in~\cite{adain}, become more similar to the original features.
This indicates that domain-invariant information can be taught within the local domain. Normalizing inputs with both global and instance statistics leads to performance improvements of $0.96\%$ and $1.73\%$, respectively, on the PACS dataset's average results.
By combining global and instance statistics, the model can benefit from both types of information, with domain-invariant information being learned from both global and local domains.
Additionally, the use of randomly mixed statistics introduces a broader diversity of domain data, further elevating performance.
We dive deeper into the impacts of the distribution range in our supplementary material.
\begin{table}[t]
\caption{
Accuracy (\%) of feature diversification variants using various mixing strategies of global and instance statistics on PACS.}
\centering
\resizebox{0.5\linewidth}{!}{
\begin{tabular}{c|c}
\toprule
Mixing components                    & Acc.  \\ \midrule
Global only: $u=0.0$                           &  81.75  \\
Instance only: $u=1.0$     & 82.52  \\
Global \& instance fixed: $u=0.5$        & 83.43  \\
\textbf{Global \& instance randomly}: $\boldsymbol{u \sim U(0,1)}$ & \textbf{84.07}     \\ \bottomrule
\end{tabular}
\label{ablation1}}
\vspace{-3mm}
\end{table}

{\noindent\bf Comparison with L2 Weight Regularization:}
As depicted in Table~\ref{ablation3}-(a), we evaluate the effects of weight regularization using the FedProx method~\cite{FedProx}, which entails L2 regularization with $\mu$ set to $0.1$.
Using only global statistics in client-agnostic learning serves as a type of regularization that acts to prevent deviation from the global model.
Notably, this method surpasses FedProx, achieving $81.75\%$ accuracy in comparison to FedProx's $80.86\%$ accuracy.
Our approach goes beyond simple L2 regularization by training the model while considering the relevance of model parameters, aiming to generate client-invariant representations.

\begin{table*}[]
\caption{Ablation studies comparing the individual components of our model, with \textbf{ours} denoted in bold.}
\centering
\resizebox{\textwidth}{!}{
\begin{tabular}{c|c|c|c|ccccc}
\toprule
& \multirow{2}{*}{Aug.} & \multirow{2}{*}{Loss} & \multirow{2}{*}{Inference} & \multicolumn{5}{c}{Acc. (\%)}                  \\ \cline{5-9} 
& & &  & P & A & C & \multicolumn{1}{c|}{S} & Avg. \\ \midrule
\multirow{2}{*}{(a) Baselines} &\multirow{2}{*}{X}    &   CE   & \multirow{2}{*}{Global statistics} & {92.74 (0.97)} & {77.08 (0.59)} & {75.08 (0.85)} & \multicolumn{1}{c|}{{78.28 (1.14)}} & {80.79} \\
    & & FedProx  & & 92.82 (0.90)         & 76.43 (0.93)          & 75.26 (0.99)          & \multicolumn{1}{c|}{78.95 (1.64)}          & 80.86 \\ \midrule
\multirow{2}{*}{(b) Effectiveness of FedFD} & Mixstyle & \multirow{2}{*}{\textbf{CAL}}  & \multirow{2}{*}{Global statistics}  & 92.04 (0.63)          & 80.19 (0.28)          & 77.14 (0.48)          & \multicolumn{1}{c|}{82.53 (0.94)}          & 82.98      \\
   & \bf{FedFD} &  & & \textbf{92.99 (0.61)} & \textbf{82.17 (1.15)} & \textbf{77.71 (1.13)} & \multicolumn{1}{c|}{\textbf{83.40 (0.19)}} & \textbf{84.07} \\ \midrule
\multirow{3}{*}{\begin{tabular}[c]{@{}c@{}}(c1) Effectiveness of \\ instance feature adapter\end{tabular}} & \multirow{3}{*}{X} & \multirow{3}{*}{CE}  & Global statistics  & {92.74 (0.97)} & {77.08 (0.59)} & {75.08 (0.85)} & \multicolumn{1}{c|}{{78.28 (1.14)}} & {80.79}   \\
                       &                    &                      & Instance statistics (TBN~\cite{nado2020evaluating})  & 15.52 (4.89) & 10.91 (2.41) & 9.36 (7.09) & \multicolumn{1}{c|}{16.09 (9.44)}  &   12.97 \\ 
                       &                    &                      & \textbf{Instance feature adapter}    & \textbf{93.11 (0.08)} & \textbf{80.93 (0.79)} & \textbf{77.01 (0.24)} & \multicolumn{1}{c|}{\textbf{82.54 (0.79)}}  & \textbf{83.40} \\ \midrule
\multirow{7}{*}{\begin{tabular}[c]{@{}c@{}}(c2) Comparison with \\ other inference methods\end{tabular}} & \multirow{7}{*}{\textbf{FedFD}} & \multirow{7}{*}{\textbf{CAL}}  & BIN~\cite{BIN}  & 93.86 (0.53) & 78.81 (0.45) & 79.54 (0.40) & \multicolumn{1}{c|}{81.94 (0.88)}  &   83.54   \\
  & &   & DSON~\cite{DSON}  & 93.70 (0.39) & 83.17 (1.09) & 76.08 (1.30) & \multicolumn{1}{c|}{83.14 (1.02)}  & 84.02 \\
  & &   & IABN~\cite{IABN}  & 93.29 (0.13) & 80.18 (1.16) & 79.38 (1.33) & \multicolumn{1}{c|}{83.58 (0.91)}  & 84.11 \\
 & &   & MixNorm~\cite{MixNorm}  & 93.60 (0.52) & 82.77 (0.41) & 78.30 (1.13) & \multicolumn{1}{c|}{83.12 (0.74)} & 84.45 \\
 & &   & Random alpha  & 90.67 (0.59) & 74.82 (2.97) & 74.96 (0.76) & \multicolumn{1}{c|}{79.03 (1.64)}  & 79.87  \\
  & &   & Fixed alpha~\cite{AlphaBN} & 93.59 (0.25) & 79.27 (1.40) & 79.64 (1.00) & \multicolumn{1}{c|}{82.43 (1.12)}  & 83.73  \\
  & &   & \textbf{Instance feature adapter}  & \textbf{94.24 (0.33)} & \textbf{84.30 (0.44)} & \textbf{79.80 (1.16)} & \multicolumn{1}{c|}{\textbf{83.79 (0.49)}}  & \textbf{85.53} \\ \bottomrule
\end{tabular}
\label{ablation3}}
\vspace{-3mm}
\end{table*}

{\noindent\bf Effectiveness of Federated Feature Diversification:}
Our FedFD stands as a feature augmentation technique specifically designed for federated DG, effectively leveraging the federated characteristics that provide server model access. This contrasts with Mixstyle~\cite{Mixstyle}, which diversifies styles through the random mixing of instance statistics within batch samples, enabling the learning of style-agnostic features.
As observed in Table~\ref{ablation3}-(b), in terms of performance across all domains, our FedFD consistently outperforms Mixstyle in a fair comparison where Mixstyle substitutes FedFD. This performance superiority stems from the utilization of global statistics, while Mixstyle is limited to single-domain data access during local training. Such findings highlight that augmenting features with global statistics can enhance the model's domain robustness.

{\noindent\bf Comparison Studies of Instance Feature Adapter:}
In Table~\ref{ablation3}-(c), we analyze the effectiveness of our instance feature adapter.
We first apply our adapter to the baseline model, SiloBN, and demonstrate its effectiveness when using instance statistics along with global statistics during testing.
In Table~\ref{ablation3}-(c1), we compare the performance of the model trained with SiloBN using global statistics for testing (conventional approach), instance statistics for testing with TBN~\cite{nado2020evaluating} holding a batch size of one, and our instance-wise approach of appropriately combining global and instance statistics during testing.
Our approach improves performance by reflecting the test domain with instance statistics on global statistics that are sensitive to distribution shift.

In the following step, we present a comparison of our adapter with various inference methods in Table~\ref{ablation3}-(c2), where we utilize FedFD as the baseline. Specifically, we compare our approach with BIN~\cite{BIN} that employs learned alpha to incorporate batch normalization (BN) and instance normalization (IN) layers in the backbone network. During training, the interpolation parameters are optimized and then applied to test samples. However, since the interpolation parameters are fitted to the training data, they can not generalize well to unseen domains.
Ensemble predictions (DSON~\cite{DSON}) from multiple local BN layers show degradation in performance on domains that are severely different from the training domains and also raise privacy concerns due to the sharing of multiple local statistics.
IABN~\cite{IABN} calibrates learned statistics with instance statistics using a rule-based function, but it requires sensitive hyper-parameters to be properly selected for each domain.
Our adapter outperforms IABN on all domains without the need for sensitive hyper-parameters.
Additionally, we compare our adapter with MixNorm~\cite{MixNorm}, which augments test input data with spatial augmentations to estimate the test distribution more accurately. While it slightly improves performance, it also increases the inference time and memory usage.
We compare with the naive approaches of random and fixed alpha~\cite{AlphaBN}. This comparison demonstrates that layer-wise and instance-wise generation strategies are markedly more effective than utilizing a fixed alpha strategy, further validating our adaptive approach.

{\noindent\bf \hspace{-0.6mm}Computational Overhead Analysis:}
Table~\ref{ablation4} presents the computational cost analysis of the various methods. The time required for training and inference was measured as the average per-iteration time with a batch size of 64 on an Intel Xeon Gold 6342 processor and a single NVIDIA RTX A5000 graphics card. FedAvg, SiloBN, and FedFD do not incur additional computational overhead during inference, while BIN requires additional processing time to calculate instance normalization.
In addition to FedFD, we also consider computation efficiency and performance of FedFD-A by incorporating inference methods such as MixNorm and IABN. MixNorm requires five times more memory and computation compared to other methods, as it forward passes multiple data (in this case, five) per batch sample. IABN calculates instance statistics at each BN layer and derives the corresponding interpolation parameter using a rule-based function, which results in better computational efficiency compared to MixNorm, but with lower performance improvement. FedFD-A achieves the best accuracy with a marginal increase in parameters and computation. Our FedFD and FedFD-A balance performance and computational overhead, offering superior efficiency and performance compared to other methods. Users can select the appropriate method based on computational requirements.

\begin{table}[t]
\caption{Comparison of computational cost and accuracy (\%) on the PACS dataset.}
\centering
\resizebox{0.47\textwidth}{!}{
\begin{tabular}{c|c|c|c|c}
\toprule
Methods & \#Parms. & Training & Inference& Acc. (\%) \\ \midrule
FedAvg~\cite{FedAvg} & 11.31M & 1.43ms & 1.34ms & 77.35\\
SiloBN~\cite{SiloBN} & 11.31M & 1.43ms & 1.34ms & 80.79\\
\textbf{FedFD} & 11.31M & 2.25ms & 1.34ms & 84.07\\ \midrule
BIN~\cite{BIN}  & 11.31M & 3.24ms & 1.35ms & 83.54\\
MixNorm~\cite{MixNorm} & 11.31M & 2.25ms & 6.70ms & 84.45\\
IABN~\cite{IABN} & 11.31M & 2.25ms & 1.57ms & 84.11\\
\textbf{FedFD-A} & 11.53M & 3.50ms & 1.37ms & 85.53\\ \bottomrule
\end{tabular}
\label{ablation4}}
\vspace{-3mm}
\end{table}

\begin{table*}[]
\caption{Comparison results on PACS, VLCS, and OfficeHome. Methods displayed in gray are associated with privacy issues.}
\centering
\resizebox{\textwidth}{!}{
\begin{tabular}{c|c|ccccc|c|c|c}
\toprule
\multirow{2}{*}{Paradigm}                                                       &  \multirow{2}{*}{Method} & \multicolumn{5}{c|}{PACS} & VLCS & OfficeHome & \multirow{2}{*}{\begin{tabular}[c]{@{}c@{}}Avg. of \\3 datasets\end{tabular}} \\ \cline{3-9} 
                                                                                &                         & P & A & C & \multicolumn{1}{c|}{S} & Avg. & Avg. & Avg. \\ \midrule
\multirow{4}{*}{\begin{tabular}[c]{@{}c@{}}Decentralized \\w/o DG\end{tabular}} & FedAvg~\cite{FedAvg}    & 90.40 (0.97) & 72.52 (2.28) & 72.59 (0.37) & \multicolumn{1}{c|}{73.90 (1.74)} & 77.35 & 74.86 & 63.47 & 71.89 \\
                                                                                & FedProx~\cite{FedProx}  & 90.49 (0.69) & 72.41 (1.06) & 73.09 (0.91) & \multicolumn{1}{c|}{72.93 (1.48)} & 77.23 & 74.42 & 63.02 & 71.56 \\ 
                                                                                & FedBN~\cite{FedBN}  & 92.63 (0.65) & 76.22 (1.05) & 72.82 (2.50) & \multicolumn{1}{c|}{76.59 (2.48)} & 79.57 & 74.85 & 63.75 & 72.72 \\
                                                                                & SiloBN~\cite{SiloBN} & \textbf{92.74 (0.97)} & \textbf{77.08 (0.59)} & \textbf{75.08 (0.85)} & \multicolumn{1}{c|}{\textbf{78.28 (1.14)}} & \textbf{80.79} & \textbf{75.33} & \textbf{64.08} & \textbf{73.40} \\ \midrule
\multirow{8}{*}{\begin{tabular}[c]{@{}c@{}}Decentralized \\w/ DG\end{tabular}} & RandAug~\cite{RandAug}  & 93.57 (0.60) & 77.15 (1.39) & 71.68 (2.14) & \multicolumn{1}{c|}{66.64 (2.34)} & 77.26 & 74.32 & \textbf{64.59} & 72.06 \\
                                                                                &                                Mixstyle~\cite{Mixstyle} & 92.75 (0.45) & \textbf{80.21 (0.33)} & \textbf{76.77 (0.72)} & \multicolumn{1}{c|}{79.98 (2.22)} & 82.43 & \textbf{75.54} & 63.24 & 73.74 \\ 
                                                                                &                                SFA~\cite{SFA}  & 87.22 (1.76) & {72.01 (3.22)} & {73.12 (1.03)} & \multicolumn{1}{c|}{79.52 (1.63)} & 77.98 & 72.77 & 59.47 & 70.06 \\ 
                                                                                &                                RandConv~\cite{RandConv}  & 91.78 (0.40) & 77.16 (1.55)          & 70.56 (3.28)          & \multicolumn{1}{c|}{78.78 (1.46)} & 79.57 & 73.68 & 63.25 & 72.17 \\ 
                                                                                &                                L2D~\cite{L2D}      & \textbf{93.59 (0.55)} & 79.69 (1.12) & 75.81 (1.37)  & \multicolumn{1}{c|}{\textbf{82.76 (0.55)}} & \textbf{82.96} & 75.37 & 63.29 & \textbf{73.87} \\ \cmidrule{2-10} 
                                                                                &                                JiGen~\cite{JiGen}     & 92.14 (0.40) & 72.51 (2.00) & 72.76 (1.24)  & \multicolumn{1}{c|}{73.34 (1.96)}  & 77.69 & 75.13 & 64.09 & 72.30 \\
                                                                                &                                RSC~\cite{RSC}       & 92.53 (0.86)          & 78.10 (0.56)          & \textbf{75.95 (1.08)}  & \multicolumn{1}{c|}{\textbf{79.82 (1.31)}}  & \textbf{81.60} & \textbf{75.90} &  63.01 & \textbf{73.50} \\
                                                                                &                                SelfReg~\cite{SelfReg}   & \textbf{93.41 (0.76)}  & \textbf{78.30 (1.16)}   & 74.94 (0.43) & \multicolumn{1}{c|}{77.13 (2.09)} & 80.94  & 72.79 & \textbf{64.85} & 72.86 \\ \midrule
\multirow{9}{*}{\begin{tabular}[c]{@{}c@{}}Decentralized \\w/ federated DG\end{tabular}}      & \cellcolor[gray]{0.9}{COPA~\cite{COPA}} & \cellcolor[gray]{0.9}94.70 (1.07) & \cellcolor[gray]{0.9}83.75 (0.29) & \cellcolor[gray]{0.9}78.58 (0.96) & \multicolumn{1}{c|}{\cellcolor[gray]{0.9}84.45 (1.33)} & \cellcolor[gray]{0.9}85.37 & \cellcolor[gray]{0.9}74.51 & \cellcolor[gray]{0.9}62.42 & \cellcolor[gray]{0.9}74.10 \\
                                                                                &  \cellcolor[gray]{0.9}{FedDG~\cite{FedDG}}  & \cellcolor[gray]{0.9}94.45 (0.44)   & \cellcolor[gray]{0.9}83.83 (0.28)    & \cellcolor[gray]{0.9}73.41 (1.33)   & \multicolumn{1}{c|}{\cellcolor[gray]{0.9}78.40 (0.73)} & \cellcolor[gray]{0.9}82.52 & \cellcolor[gray]{0.9}75.28 & \cellcolor[gray]{0.9}64.90 & \cellcolor[gray]{0.9}74.23\\ 
                                                                                &  \cellcolor[gray]{0.9}{CCST~\cite{CCST}}  & \cellcolor[gray]{0.9}93.59 (0.49)   & \cellcolor[gray]{0.9}80.22 (0.55)    & \cellcolor[gray]{0.9}77.79 (1.95)   & \multicolumn{1}{c|}{\cellcolor[gray]{0.9}82.61 (0.84)} & \cellcolor[gray]{0.9}83.56 & \cellcolor[gray]{0.9}75.20 & \cellcolor[gray]{0.9}63.50 & \cellcolor[gray]{0.9}74.09\\ \cmidrule{2-10} 
                                                                                &           CSAC~\cite{CSAC} & 94.83 (0.42) & 80.48 (1.25)  & 75.46 (1.58) & \multicolumn{1}{c|}{{79.56 (1.35)}} & {82.58} & 75.21 & 64.35 & 74.05 \\ 
                                                                                &           FedSR~\cite{FedSR} & 91.71 (0.62) & 76.49 (1.26) & 72.87 (1.56) & \multicolumn{1}{c|}{{75.46 (2.22)}} & {79.13} & 75.36 & 63.45 & 72.65 \\ 
                                                                                &           FedSAM~\cite{FedSAM} & 90.73 (0.89) & 74.13 (1.96)  & 73.46 (0.99) & \multicolumn{1}{c|}{{75.97 (1.39)}} & {78.58} & 74.37 & 61.93 & 71.63 \\ 
                                                                                &           SiloBN w/ GA~\cite{GA} & \textbf{94.91 (0.59)} & 78.10 (1.42) & 71.15 (0.67) & \multicolumn{1}{c|}{{79.17 (0.50)}} & {80.83} & 75.41 & 64.19 & 73.48 \\ 
                                                                                &           \textbf{FedFD}    & 92.99 (0.61)  & 82.17 (1.15)  & 77.71 (1.13)  & \multicolumn{1}{c|}{83.40 (0.19)} & 84.07 & 76.62 & 64.01 & 74.90 \\ 
                                                                                &           \textbf{FedFD-A} & 94.24 (0.33)  & \textbf{84.30 (0.44)}  & \textbf{79.80 (1.16)}  & \multicolumn{1}{c|}{\textbf{83.79 (0.49)}} & \textbf{85.53} & \textbf{76.68} & \textbf{64.71} & \textbf{75.64} \\
                                                                                \bottomrule
\end{tabular}
}
\label{main_table1}
\end{table*}

\subsection{Comparison with Competitive Methods}
{\noindent\bf Competitive Methods:}
In Table~\ref{main_table1}, we compare our FedFD-A with representative methods of FL and DG in the classification task. (1) Federated learning methods: FedAvg~\cite{FedAvg}, FedProx~\cite{FedProx}, FedBN~\cite{FedBN}, and SiloBN~\cite{SiloBN}; (2) A data augmentation method: RandAug~\cite{RandAug}; (3) Augmentation-based DG methods: Mixstyle~\cite{Mixstyle}, SFA~\cite{SFA}, RandConv~\cite{RandConv}, and L2D~\cite{L2D}; (4) Regularization-based DG methods: JiGen~\cite{JiGen}, RSC~\cite{RSC}, and SelfReg~\cite{SelfReg}; (5) Federated DG methods: COPA~\cite{COPA}, FedDG~\cite{FedDG}, CCST~\cite{CCST}, CSAC~\cite{CSAC}, FedSR~\cite{FedSR}, FedSAM~\cite{FedSAM}, and GA~\cite{GA}.

{\noindent\bf Performance Analysis:}
Our analysis focuses on the results obtained from PACS, where the domain distribution shift is substantial.
{
It is observed that decentralized methods without DG fail to provide high performance compared to other approaches. While FedProx regularizes the local models, it fails to address the domain shift. FedBN and SiloBN mitigate the domain shift through the use of local BN layers, but they are limited in their performance as they do not tackle the domain shift that occurs between training and test distributions.
}
The decentralized methods with DG demonstrate improved performance across several domains. In particular, augmentation-based DG methods achieve substantial accuracy improvements, although there are some domains where performance is significantly degraded. This is due to the fact that such methods only learn domain-invariant representations within a single domain and are not effective if the augmentation applied to the source domain does not cover the test distribution.
JiGen, RSC, and SelfReg consistently improve or maintain the performance on four domains compared to FedAvg, but the improvement is very marginal.
While COPA, FedDG, and CCST achieve competitive performance across multiple domains by enabling the model to learn client-invariant representations from other clients, they do so at the expense of serious privacy concerns. On the other hand, our methods, FedFD and FedFD-A, deliver state-of-the-art performance without compromising client privacy. It is noteworthy that amongst the methods that preserve privacy, ours exhibits superior performance.
On VLCS and OfficeHome, where the domain shift is relatively small, FedFD and FedFD-A consistently improve the performance on almost all domains compared to other methods, demonstrating their robustness across a wide range of domains. Further results on in-domain performance and a cross-silo FL setting can be found in supplementary material.

\section{Discussion}
In this study, we assume the utilization of BN layers in local models, an assumption that may not always be applicable. Our objective is to leverage both local and global characteristics during the local models' training stage. The decision to employ BN layers stems from their capacity to encapsulate domain-specific information.
Even in the absence of BN layers in certain models, there exist components that learn domain-specific information, such as parts of the model or hypernetworks~\cite{sun2021partialfed, DoPrompt, shamsian2021personalized}. We can apply random interpolation to these components between the local and global models.

We have tested the effectiveness of our method on Transformer-based multi-source domain generalization with the use of DoPrompt~\cite{DoPrompt}. DoPrompt introduces domain-specific prompts to the input during training to capture domain-related information. Each domain prompt is optimized for a single domain, enabling efficient learning of domain-specific knowledge. Although DoPrompt includes a prompt adapter that generates appropriate prompts for each input image based on the learned domain prompts, we chose not to utilize the prompt adapter in our demonstrations to focus on the efficacy of our federated diversification approach.
We adapt DoPrompt to a federated learning setting with domain prompts placed locally, akin to SiloBN~\cite{SiloBN} and FedBN~\cite{FedBN}, and employ our FedFD algorithm using local and global domain prompts to create interpolated prompts. We then train the local models with our proposed loss function and evaluate our final aggregated global model on unseen domains.
We show the results of applying DoPrompt in federated learning, with and without GA~\cite{GA}, and our method in Table~\ref{doprompt}.
The experiment shows that our method extends to models with various domain-specific components, moving beyond those that rely solely on batch normalization.

\begin{table}[t]
\centering
\caption{Comparison results on PACS and VLCS using the transformer architecture.}
\begin{subtable}[h]{0.49\textwidth}
\centering
\subcaption{Performance (\%) on PACS}
\resizebox{1.0\textwidth}{!}{
\begin{tabular}{c|ccccc}
\toprule                                                      
\multirow{2}{*}{Method} & \multicolumn{5}{c}{PACS} \\ \cline{2-6} 
                        & P & A & C & \multicolumn{1}{c|}{S} & Avg. \\ \midrule
DoPrompt~\cite{DoPrompt} & 99.24 (0.16) & 90.56 (0.54) & 81.06 (0.42) & \multicolumn{1}{c|}{77.71 (0.56)} & 87.14 \\ 
{\begin{tabular}[c]{@{}c@{}}DoPrompt \\w/ GA~\cite{GA}\end{tabular}} & 99.33 (0.24) & 89.98 (0.53) & 81.03 (0.24) & \multicolumn{1}{c|}{79.01 (1.21)} & 87.34 \\ 
\textbf{FedFD-A} & \textbf{99.39 (0.06)} & \textbf{92.29 (0.42)} & \textbf{83.88 (1.07)} & \multicolumn{1}{c|}{\textbf{79.62 (1.23)}} & \textbf{88.79} \\ 
\bottomrule
\end{tabular}}
\end{subtable}%
\hfill
\begin{subtable}[h]{0.49\textwidth}
\subcaption{Performance (\%) on VLCS}
\centering
\resizebox{1.0\textwidth}{!}{
\begin{tabular}{c|ccccc}
\toprule                                                      
\multirow{2}{*}{Method} & \multicolumn{5}{c}{VLCS} \\ \cline{2-6} 
                        & V & L & C & \multicolumn{1}{c|}{S} & Avg. \\ \midrule
DoPrompt~\cite{DoPrompt} & 78.32 (1.29) & \textbf{65.94 (1.28)} & 97.02 (0.49) & \multicolumn{1}{c|}{77.80 (1.22)} & 79.82 \\ 
{\begin{tabular}[c]{@{}c@{}}DoPrompt \\w/ GA~\cite{GA}\end{tabular}} & 79.84 (0.49) & 65.15 (0.81) & 97.14 (0.17) & \multicolumn{1}{c|}{76.72 (0.59)} & 79.71 \\ 
\textbf{FedFD-A} & \textbf{81.41 (0.81)} & 65.06 (0.62) & \textbf{97.46 (0.79)} & \multicolumn{1}{c|}{\textbf{79.64 (0.94)}} & \textbf{80.89} \\ 
\bottomrule
\end{tabular}}
\end{subtable}
\label{doprompt}
\end{table}

\section{Conclusion}
We propose two methods specifically designed for federated domain generalization. The first method suggests augmenting local data into various domains by utilizing the global model at the server. This technique encourages each local model to learn client-invariant representations.
To improve the model's performance on unseen domains after training, we propose utilizing an instance adaptation method. This approach utilizes the test sample's instance statistics dynamically during inference when a test sample is presented.
Our methods were rigorously verified through a series of experiments on domain generalization benchmarks. The results effectively demonstrated their applicability and efficiency in a federated learning environment.

%
%
\bibliographystyle{splncs04}
\bibliography{main}

\clearpage

\title{Supplemantary Material on Feature Diversification and Adaptation for Federated Domain Generalization} 


\author{Seunghan Yang\orcidlink{0000-0002-0411-8407} \and
Seokeon Choi\orcidlink{0000-0002-1695-5894} \and
Hyunsin Park\orcidlink{0000-0003-3556-5792} \and
Sungha Choi\orcidlink{0000-0003-2313-9243} \and \\
Simyung Chang\orcidlink{0000-0001-7750-191X} \and
Sungrack Yun\orcidlink{0000-0003-2462-3854}}

\authorrunning{S. Yang et al.}
\titlerunning{Feature Diversification and Adaptation for Federated Domain Generalization}
\institute{Qualcomm AI Research$^{\dagger}$ \\
\email{\{seunghan,seokchoi,hyunsinp,sunghac,simychan,sungrack\}\\@qti.qualcomm.com}}

\maketitle

\def\thesection{\Alph{section}}
\setcounter{section}{0}

\section{Experimental Details}
{\let\thefootnote\relax\footnotetext{{
\hspace{-3mm}$\dagger$ Qualcomm AI Research is an initiative of Qualcomm Technologies, Inc.}}}
Our framework is built upon CSAC\footnote{https://github.com/junkunyuan/csac}~\cite{CSAC}, and we follow the experimental setup of DomainBed\footnote{https://github.com/facebookresearch/DomainBed}~\cite{Domainbed}.
As described in the main paper, we use the training-domain validation set for model selection.
To ensure fair comparison, we implement and evaluate all competitive methods using their publicly available codes.
\textbf{It should be noted that our evaluation protocol may produce results that differ from those reported in the original papers, which utilized the test-domain validation set for model selection.}

\subsection{Experimental Results on VLCS and OfficeHome}
In Table~\ref{vlcs_officehome}, our proposed FedFD and FedFD-A methods achieve state-of-the-art performance on the VLCS~\cite{VLCS} benchmark. In contrast, several methods exhibit decreased performance on the VOC2007 (V) or SUN (S) domains compared to the FedAvg~\cite{FedAvg} baseline.
Regarding the OfficeHome~\cite{OfficeHome} benchmark, we show that most methods do not exhibit significant improvement due to the small domain shift across clients.
Our FedFD-A shows performance improvement in most domains even in situations with minimal domain shift, and it demonstrates high average performance.

\setcounter{table}{6}
\begin{table*}[h]
\caption{Comparison results on VLCS and OfficeHome. Methods displayed in gray are associated with privacy issues.}
\centering
\begin{subtable}{0.5\textwidth}
\centering
\resizebox{0.95\textwidth}{!}{
\centering
\begin{tabular}{c|cccccc}
\toprule
\multirow{2}{*}{Method} & \multicolumn{5}{c}{VLCS}                  \\ \cline{2-6}
                        & V & L & C & \multicolumn{1}{c|}{S} & Avg. \\ \midrule
FedAvg~\cite{FedAvg}    & \textbf{73.53 (0.87)} & 57.92 (1.08) & 96.22 (0.93) & \multicolumn{1}{c|}{71.77 (1.15)} & 74.86 \\
FedProx~\cite{FedProx}  & 73.04 (0.53) & 57.85 (1.27) & 95.97 (0.75) & \multicolumn{1}{c|}{70.83 (3.91)} & 74.42\\
FedBN~\cite{FedBN}      & 71.50 (0.71) & 58.65 (0.70) & \textbf{96.86 (0.29)} & \multicolumn{1}{c|}{72.39 (2.45)} & 74.85\\
SiloBN~\cite{SiloBN}    & 71.78 (0.98) & \textbf{58.71 (1.14)} & 96.67 (0.53) & \multicolumn{1}{c|}{\textbf{74.15 (1.59)}} & \textbf{75.33} \\ \midrule
RandAug~\cite{RandAug}  & 72.42 (0.78) & 58.32 (0.74) & 95.79 (0.86) & \multicolumn{1}{c|}{70.76 (2.27)} & 74.32 \\
Mixstyle~\cite{Mixstyle}& 72.61 (0.66) & 58.52 (0.66) & 97.69 (0.51) & \multicolumn{1}{c|}{\textbf{73.33 (1.37)}} & \textbf{75.54} \\ 
SFA~\cite{SFA}          & 65.08 (0.81) & \textbf{61.55 (1.22)} & 96.29 (1.19) & \multicolumn{1}{c|}{68.15 (1.07)} & 72.77 \\ 
RandConv~\cite{RandConv}& 70.83 (0.90) & 57.16 (1.50) & 95.64 (0.23) & \multicolumn{1}{c|}{71.08 (2.24)} & 73.68 \\ 
L2D~\cite{L2D}          & \textbf{72.84 (1.59)} & 59.78 (0.69) & \textbf{97.97 (0.54)} & \multicolumn{1}{c|}{70.89 (1.83)} & 75.37 \\ \cmidrule{1-6} 
JiGen~\cite{JiGen}      & 73.65 (0.57)  & 58.09 (0.69) &  \textbf{98.06 (0.35)} & \multicolumn{1}{c|}{70.72 (1.21)}  &  75.13    \\
RSC~\cite{RSC}          & \textbf{75.27 (1.00)} & 59.79 (1.22)  & 97.01 (1.01) & \multicolumn{1}{c|}{71.51 (1.00)}  &  \textbf{75.90}    \\
SelfReg~\cite{SelfReg}  & 68.13 (0.76)  & \textbf{60.37 (0.66)}  & 88.64 (3.09)  & \multicolumn{1}{c|}{\textbf{74.03 (0.19)}}  & 72.79 \\ \midrule
\cellcolor[gray]{0.9}{COPA~\cite{COPA}}                 & \cellcolor[gray]{0.9}71.50 (1.05) & \cellcolor[gray]{0.9}61.00 (0.89) & \cellcolor[gray]{0.9}93.83 (0.41) & \multicolumn{1}{c|}{\cellcolor[gray]{0.9}71.72 (0.74)} & \cellcolor[gray]{0.9}74.51  \\
\cellcolor[gray]{0.9}{FedDG~\cite{FedDG}}  & \cellcolor[gray]{0.9}71.05 (0.62) & \cellcolor[gray]{0.9}59.46 (1.08) & \cellcolor[gray]{0.9}96.64 (0.86) & \multicolumn{1}{c|}{\cellcolor[gray]{0.9}73.96 (0.74)} & \cellcolor[gray]{0.9}75.28 \\ 
\cellcolor[gray]{0.9}{CCST~\cite{CCST}}  & \cellcolor[gray]{0.9}72.19 (0.78) & \cellcolor[gray]{0.9}59.34 (0.63) & \cellcolor[gray]{0.9}97.01 (0.50) & \multicolumn{1}{c|}{\cellcolor[gray]{0.9}72.24 (0.86)} & \cellcolor[gray]{0.9}75.20 \\ \cmidrule{1-6} 
CSAC~\cite{CSAC}       & 72.96 (0.91) & \textbf{59.78 (0.84)} & 96.52 (0.38) & \multicolumn{1}{c|}{71.60 (1.20)} & 75.21  \\ 
FedSR~\cite{FedSR}     & \textbf{74.04 (0.97)} & 58.20 (0.47) & 97.39 (0.64) & \multicolumn{1}{c|}{71.83 (0.92)} & 75.36  \\ 
FedSAM~\cite{FedSAM}   & 72.08 (0.58) & 58.04 (1.98) & 95.80 (0.79) & \multicolumn{1}{c|}{71.57 (1.64)} & 74.37  \\ 
SiloBN w/ GA~\cite{GA} & 72.53 (0.98) & 58.03 (0.42) & 96.89 (0.45) & \multicolumn{1}{c|}{74.20 (1.07)} & 75.41  \\ 
\textbf{FedFD}   & 73.69 (0.90) & 59.14 (0.74) & \textbf{98.00 (0.29)} & \multicolumn{1}{c|}{\textbf{75.65 (1.00)}} & 76.62 \\ 
\textbf{FedFD-A} & 73.93 (1.14) & 59.29 (0.28) & 97.84 (0.37) & \multicolumn{1}{c|}{75.64 (0.86)} & \textbf{76.68} \\ \bottomrule
\end{tabular}}
\end{subtable}%
\hfill
\begin{subtable}{0.5\textwidth}
\centering
\resizebox{0.95\textwidth}{!}{
\centering
\begin{tabular}{c|cccccc}
\toprule
\multirow{2}{*}{Method} & \multicolumn{5}{c}{OfficeHome}                  \\ \cline{2-6} 
                        & A & C & P & \multicolumn{1}{c|}{R} & Avg. \\ \midrule
FedAvg~\cite{FedAvg}                  & 56.82 (0.31) & 50.53 (0.60) & 72.38 (0.24) & \multicolumn{1}{c|}{\textbf{74.16 (0.38)}} & 63.47 \\
FedProx~\cite{FedProx}                & 56.58 (0.63) & 49.44 (0.39) & 72.15 (0.30) & \multicolumn{1}{c|}{73.90 (0.46)} & 63.02 \\
FedBN~\cite{FedBN}                 & \textbf{58.31 (0.36)} & 50.07 (0.49) & 72.53 (0.70) & \multicolumn{1}{c|}{74.06 (0.50))} & 63.75 \\
SiloBN~\cite{SiloBN}         & 58.29 (0.70) & \textbf{51.16 (0.48)} & \textbf{72.80 (0.59)} & \multicolumn{1}{c|}{74.06 (0.51)} & \textbf{64.08} \\ \midrule
RandAug~\cite{RandAug}    & \textbf{58.50 (0.34)} & 52.18 (0.77) & \textbf{73.17 (0.36)} & \multicolumn{1}{c|}{\textbf{74.52 (0.52)}} & \textbf{64.59} \\
Mixstyle~\cite{Mixstyle}                & 55.57 (1.24) & 53.31 (0.65) & 70.90 (0.81) & \multicolumn{1}{c|}{73.18 (0.27)} & 63.24 \\ 
SFA~\cite{SFA}                 & 50.99 (1.20) & 50.97 (0.35) & 66.84 (0.65) & \multicolumn{1}{c|}{69.08 (0.23)} & 59.47 \\ 
RandConv~\cite{RandConv}             & 56.19 (0.80) & 53.20 (0.47) & 71.66 (0.57) & \multicolumn{1}{c|}{71.94 (0.46)} & 63.25 \\ 
L2D~\cite{L2D}       & 54.70 (1.43) & \textbf{56.36 (0.28)} & 69.96 (0.85) & \multicolumn{1}{c|}{72.13 (0.35)} & 63.29 \\ \cmidrule{1-6} 
JiGen~\cite{JiGen}                   & 58.20 (0.67)  &  50.00 (0.02) & \textbf{73.99 (0.56)}  & \multicolumn{1}{c|}{74.18 (0.00)}  &    64.09  \\
RSC~\cite{RSC}                   & 56.67 (0.59)  & 49.59 (1.05)  & 71.61 (0.47)  & \multicolumn{1}{c|}{74.16 (0.49)}  & 63.01 \\
SelfReg~\cite{SelfReg}                     &  \textbf{59.26 (0.26)} & \textbf{51.84 (0.86)} & 73.46 (0.18)  & \multicolumn{1}{c|}{\textbf{74.85 (0.62)}}  & \textbf{64.85} \\ \midrule
\cellcolor[gray]{0.9}{COPA~\cite{COPA}}                 & \cellcolor[gray]{0.9}53.39 (0.16) & \cellcolor[gray]{0.9}57.46 (0.47) & \cellcolor[gray]{0.9}68.61 (0.15) & \multicolumn{1}{c|}{\cellcolor[gray]{0.9}70.24 (0.38)} & \cellcolor[gray]{0.9}62.42  \\
\cellcolor[gray]{0.9}{FedDG~\cite{FedDG}}  & \cellcolor[gray]{0.9}59.87 (0.06) & \cellcolor[gray]{0.9}53.51 (0.31) & \cellcolor[gray]{0.9}72.81 (0.89) & \multicolumn{1}{c|}{\cellcolor[gray]{0.9}73.41 (0.41)} & \cellcolor[gray]{0.9}64.90 \\ 
\cellcolor[gray]{0.9}{CCST~\cite{CCST}}  & \cellcolor[gray]{0.9}56.65 (0.84) & \cellcolor[gray]{0.9}25.60 (0.22) & \cellcolor[gray]{0.9}71.80 (0.53) & \multicolumn{1}{c|}{\cellcolor[gray]{0.9}72.96 (0.36)} & \cellcolor[gray]{0.9}63.50 \\ \cmidrule{1-6} 
CSAC~\cite{CSAC}         & \textbf{58.97 (1.13)} & 51.61 (0.26) & \textbf{72.57 (0.18)} & \multicolumn{1}{c|}{74.25 (0.49)} & 64.35  \\ 
FedSR~\cite{FedSR}       & 57.59 (0.49) & 49.62 (0.91) & 72.24 (0.64) & \multicolumn{1}{c|}{74.36 (0.18)} & 63.45  \\ 
FedSAM~\cite{FedSAM}     & 55.25 (0.48) & 49.54 (0.57) & 71.07 (0.14) & \multicolumn{1}{c|}{72.77 (0.36)} & 62.16  \\ 
SiloBN w/ GA~\cite{GA}   & 58.93 (1.13) & 51.06 (0.41) & 72.42 (0.87) & \multicolumn{1}{c|}{74.36 (0.11)} & 64.19  \\ 
\textbf{FedFD}   &  57.68 (0.64) & 52.90 (0.13) & 71.79 (0.47) & \multicolumn{1}{c|}{73.68 (0.39)} & 64.01 \\ 
\textbf{FedFD-A}  &  58.62 (0.51) & \textbf{53.47 (0.21)} & 72.34 (0.45) & \multicolumn{1}{c|}{\textbf{74.39 (0.17)}} & \textbf{64.71} \\ \bottomrule
\end{tabular}}
\end{subtable}
\label{vlcs_officehome}
\end{table*}
\begin{table*}[t]
\caption{Comparison results on DomainNet.}
\centering
\resizebox{0.65\textwidth}{!}{
\begin{tabular}{c|cccccccc}
\toprule
\multirow{2}{*}{Method} & \multicolumn{7}{c}{DomainNet}                  \\ \cline{2-8}
                        & C & I & P & Q & R & \multicolumn{1}{c|}{S} & Avg. \\ \midrule
SiloBN~\cite{SiloBN}    & 70.94 (1.32) & 34.87 (1.73) & 59.88 (1.34) & 58.33 (1.47) & 67.22 (1.24) & \multicolumn{1}{c|}{60.90 (0.75)} & 58.69 \\ 
SiloBN w/ GA~\cite{GA}  & \textbf{72.84 (1.68)} & 34.61 (0.33) & 60.92 (0.41) & 60.10 (1.80) & 66.66 (1.63) & \multicolumn{1}{c|}{61.58 (1.04)} & 59.45 \\ 
\textbf{FedFD-A}          & 71.38 (0.90) & \textbf{35.39 (1.26)} & \textbf{61.48 (1.22)} & \textbf{60.21 (1.71)} & \textbf{69.12 (1.28)} & \multicolumn{1}{c|}{\textbf{64.71 (2.03)}} & \textbf{60.38} \\ \bottomrule
\end{tabular}
}
\label{domainnet}
\end{table*}

\subsection{Experimental Results on DomainNet}
DomainNet~\cite{DomainNet} comprises natural images sourced from six distinct domains: Clipart (C), Infograph (I), Painting (P), Quickdraw (Q), Real (R), and Sketch (S). Data on each client originates exclusively from one domain, resulting in the domain shift across clients with different data sources. Consistent with the approaches in~\cite{FedBN, GA}, we isolate the top ten most prevalent classes to construct a subset for our experimentation. Our adopted architecture is AlexNet modified to include batch normalization (BN) after each convolutional and fully-connected layer.
For training baselines and our approach, all hyper-parameters remain consistent with those detailed in the primary experiments of the main paper conducted on PACS and VLCS datasets.
As shown in Table~\ref{domainnet}, our method consistently outperforms the baseline, SiloBN~\cite{SiloBN}. Furthermore, it exhibits superior performance compared to the recent state-of-the-art method, GA~\cite{GA}.

\setcounter{figure}{3}
\begin{figure*}[t]
\centering
\epsfig{figure=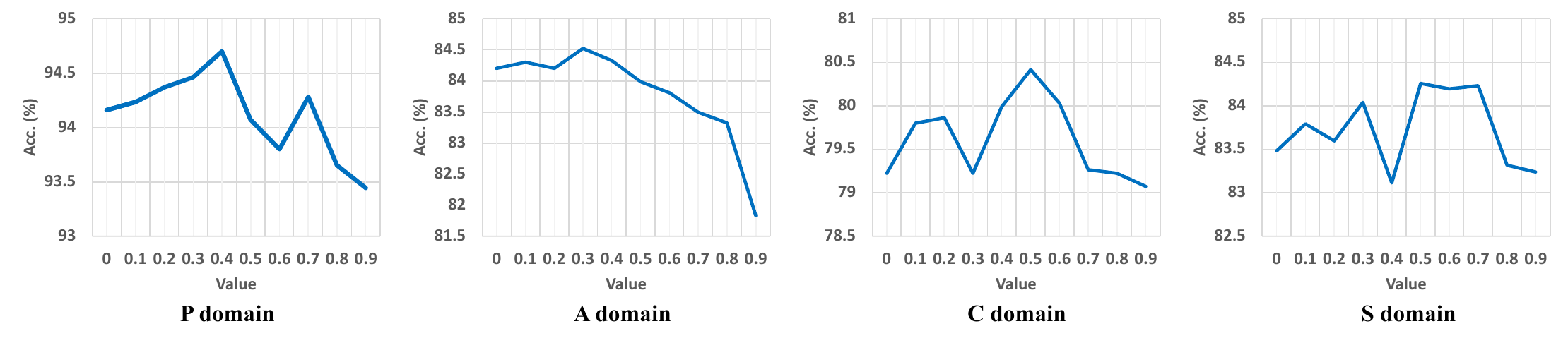,width=1.0\linewidth}
	\caption{
	Ablation studies of client-agnostic classification loss on PACS.
	We conduct experiments using various $\lambda_{1}$ (x-axis) and get the performance (y-axis).
	}
 \label{loss_fig1}
 \vspace{-2mm}
\end{figure*}

\begin{figure*}[t]
\centering
\epsfig{figure=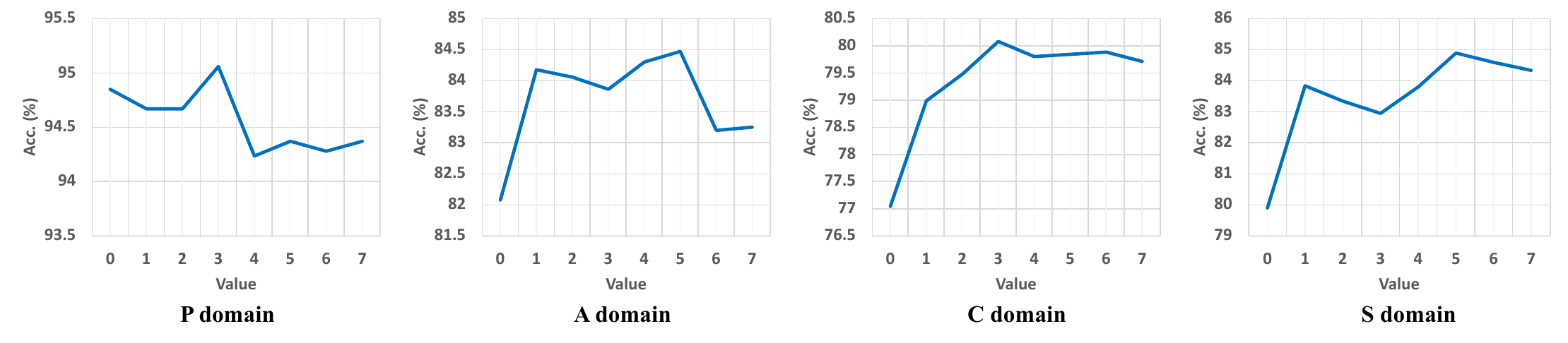,width=1.0\linewidth}
	\caption{
	Ablation studies of client-agnostic feature loss on PACS.
	We conduct experiments using various $\lambda_{2}$ (x-axis) and get the performance (y-axis).
	} 
 \label{loss_fig2}
 \vspace{-2mm}
\end{figure*}

\section{Analysis on Hyper-parameters}
\subsection{Balancing Parameter of Loss Functions}
Our objectives involve two hyper-parameters, namely $\lambda_{1}$ and $\lambda_{2}$, and we investigate the performance sensitivity of these parameters. This is a crucial ablation study in domain generalization, where the model must generalize well across several domains without depending sensitively on hyper-parameters. In Fig.~\ref{loss_fig1} and~\ref{loss_fig2}, we conduct experiments by varying each parameter while holding the other constant, specifically $\lambda_{1} = 0.1$ and $\lambda_{2} = 4.0$. Although the optimal hyper-parameters vary with each domain, choosing $\lambda_{1}$ in the range of $[0.1, 0.5]$ and $\lambda_{2}$ in the range of $[2.0, 5.0]$ yields consistent results across all domains. Within this range, FedFD-A consistently achieves high performance on all PACS domains compared to competitive methods (refer to Table 4 in the main paper). While optimal hyper-parameters can be selected for each domain, we set $\lambda_{1}$ to $0.1$ and $\lambda_{2}$ to $4.0$ across all datasets for our experiments.

\subsection{Sampling Function for Feature Diversification}
Table~\ref{extrapolation} presents the results of our experiments on the effect of using different ranges of uniform distribution for mixing instance and global statistics of Eq. 2 in the main paper. In contrast to previous augmentation-based DG methods, our proposed method interpolates local and global information, resulting in consistently improved performance across all domains.
We can use safe and diverse augmentation with the information of local domains shared by the global model.
We observe that the optimal range of augmentation differs for each test domain, indicating that the strength and type of augmentation desired for each domain vary. Specifically, with extrapolation using $U(-0.1,1.1)$, we achieve an improvement in performance on the Sketch (S) domain, which is severely deviated from Photo (P), Art painting (A), and Cartoon (C) domains.
Employing extrapolated statistics results in a greater diversity of samples compared to utilizing interpolated statistics; however, this approach causes a decline in performance on domains that are not severely different. In all our experiments, we apply the uniform distribution $U(0,1)$ across all benchmarks without selecting specific hyper-parameters for each dataset.

\begin{table}[h]
\caption{Accuracy (\%) on PACS and VLCS using various range of distribution for $u^l$ in federated feature diversification (FedFD).}
\centering
\resizebox{0.5\textwidth}{!}{
\begin{tabular}{c|ccccc|c}
\toprule                                                      
\multirow{2}{*}{Distribution} & \multicolumn{5}{c|}{PACS} & VLCS \\ \cline{2-7} 
                        & P & A & C & \multicolumn{1}{c|}{S} & Avg. & Avg. \\ \midrule
$U(0,1)$ & 94.24 (0.33) & 84.30 (0.44) & 79.80 (1.16) & \multicolumn{1}{c|}{83.79 (0.49)} & 85.53 & {\bf 76.68} \\ 
$U(0.0,0.5)$ & 94.46 (0.21) & {\bf 84.50 (0.17)} & {\bf 80.97 (1.15)} & \multicolumn{1}{c|}{84.46 (0.09)} & {\bf 86.10} & 76.42 \\ 
$U(0.5,1.0)$ & {\bf 94.58 (0.21)} & 82.89 (0.59) & 78.14 (1.48) & \multicolumn{1}{c|}{83.18 (0.04)} & 84.69 & 76.36 \\
$U(-0.1,1.1)$ & 94.46 (0.64) & 83.64 (0.83) & 80.12 (0.36) & \multicolumn{1}{c|}{\bf 84.59 (0.05)} & 85.70 & 76.46 \\ 
\bottomrule
\end{tabular}}
\label{extrapolation}
\end{table}

\begin{figure}[h]
\centering
\epsfig{figure=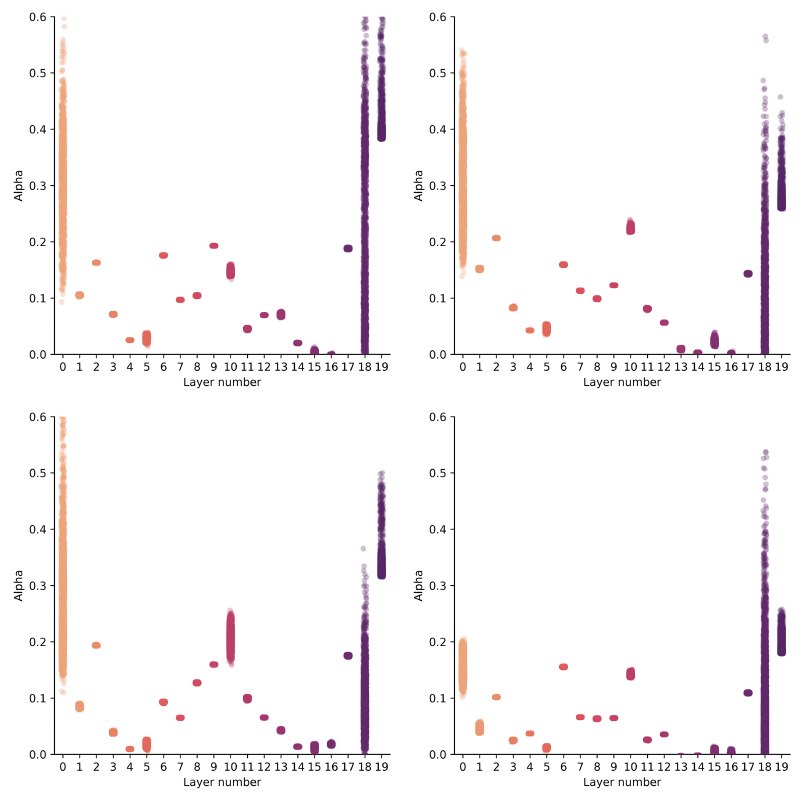,width=0.7\linewidth}
	\caption{
	Interpolation (alpha) values for all test samples on each layer are plotted for the P (upper left), A (upper right), C (lower left), and S (lower right) domains.
	}
 \label{alpha}
\end{figure}

\section{Analysis of Instance Feature Adaptation}
At test time, the interpolation values obtained from the trained adapter are utilized for inference. Interpolation values on each layer from test samples in P, A, C, and S domains are plotted in Fig.~\ref{alpha}. In PACS, distribution of interpolation values on each layer is similar in P, A, and C domains. Different interpolation values are used for test samples between $0.1$ and $0.6$ in the low-level layers, indicating the use of varying amounts of instance statistics. In S domain, interpolation values at the low-level layers are lower than other domains, which could be attributed to the large domain gap between S and other domains.
This observation aligns with the well-known concept that the use of instance normalization at lower-level features can reduce domain-specific information.
The interpolation values in the middle-level layers are almost the same across test samples, such as almost all test samples obtaining $0.2$ on the $9$-th BN layer in P domain.
At the high-level layers, which aim to generate discriminative features for classification, there is a slight variation in the interpolation values between domains.

\section{Additional Experiments}

\begin{figure}[t]
\centering
\begin{subfigure}{0.5\textwidth}
\vspace{-3mm}
\centering
\epsfig{figure=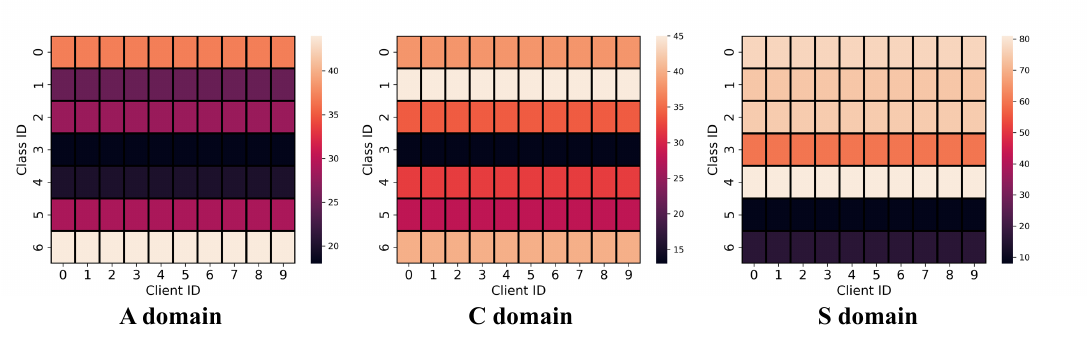,width=1.0\linewidth}
 \caption{}
 \end{subfigure}
\vspace{-0.5mm}
\begin{subfigure}{0.5\textwidth}
\centering
\epsfig{figure=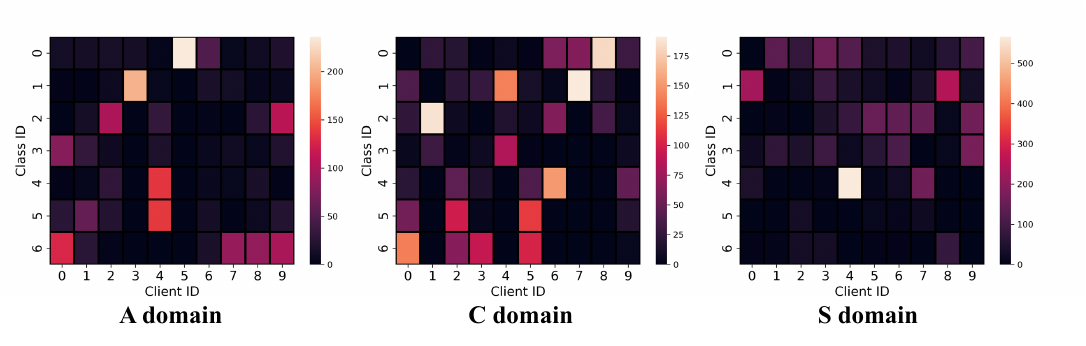,width=1.0\linewidth}
 \caption{}
 \end{subfigure}
\vspace{-0.5mm}
 \begin{subfigure}{0.5\textwidth}
\centering
\epsfig{figure=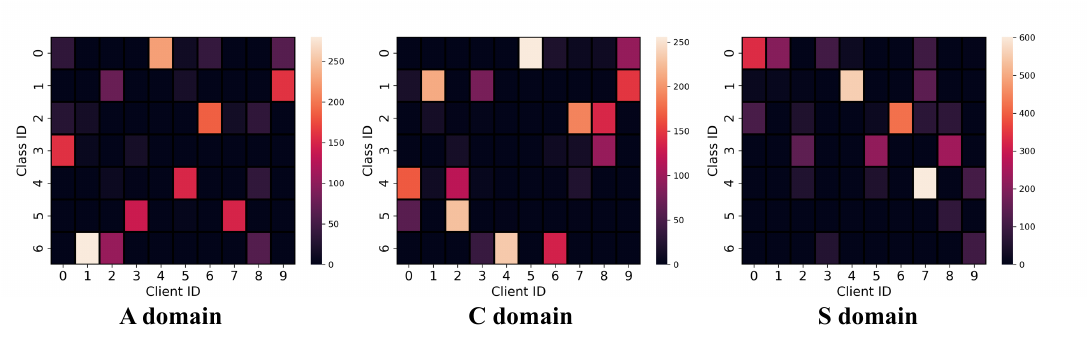,width=1.0\linewidth}
 \caption{}
 \end{subfigure}
	\caption{The data distribution is shown for (a) iid data partition, (b) non-iid data partition following Dirichlet distribution with $\alpha=0.5$, and (c) non-iid data partition following Dirichlet distribution with $\alpha=0.1$. The color bar indicates the number of data samples, while the x-axis indicates the client ID and the y-axis indicates the class ID. Each rectangle corresponds to the number of data samples for a specific class in a client. We have a total of 30 clients participating in the federated learning, with each client having data from one of the A, C, and S domains, and we evaluate the performance of the federated learning model on the P domain.
}
\label{class_client_distribution}
\vspace{-3mm}
\end{figure}

\subsection{Cross-silo with Non-IID Label Distribution}
\noindent{\bf Experimental Setup:}
We expand the number of training clients from 3 to 30 on PACS. To this end, we allocate the dataset among 10 clients, resulting in a limited quantity of data per client. We explore both iid and non-iid label distributions among clients. For iid label distribution, data is evenly divided among the 10 clients, while for non-iid label distribution, we employ a Dirichlet distribution with parameters $\alpha=0.5$ and $\alpha=0.1$. The client-specific class distribution is depicted in Figure~\ref{class_client_distribution}. During local training, 10 clients are randomly selected in each round, culminating in 120 rounds with 20 iterations per round.

\noindent{\bf Experimental Results:} Tables~\ref{30trainingclients_1}, \ref{30trainingclients_2}, and \ref{30trainingclients_3} provide a comparative analysis of our proposed FedFD-A method against FedProx~\cite{FedProx} and SiloBN~\cite{SiloBN}. As detailed in Table~\ref{30trainingclients_1}, our methodology is highly effective when dealing with a larger number of clients, showing its ability to train the global model that successfully improves generalization across various scenarios.

In our experiments concerning non-iid label distribution, although FedFD-A exhibits a slight decrease in performance as opposed to the iid scenario, we believe that the negative impact can be diminished by implementing a suitable strategy tailored for non-iid label conditions—a prospect for our future research. Despite not having been explicitly designed to accommodate non-iid label distribution, our approach nonetheless addresses the embedded domain shift challenges, as evidenced in Table~\ref{30trainingclients_2} and~\ref{30trainingclients_3}. These results affirm that our proposed FedFD-A can effectively manage the domain shift problem, even in the presence of significant label shift.

\begin{table}[t]
\centering
\caption{Comparison results on PACS in the cross-silo setup.}
\vspace{-3mm}
\begin{subtable}[h]{0.55\textwidth}
\centering
\subcaption{Performance (\%) on iid label distribution}
\vspace{-3mm}
\resizebox{1.0\textwidth}{!}{
\begin{tabular}{c|ccccc}
\toprule                                                      
\multirow{2}{*}{Method} & \multicolumn{5}{c}{PACS} \\ \cline{2-6} 
                        & P & A & C & \multicolumn{1}{c|}{S} & Avg. \\ \midrule
FedProx~\cite{FedProx} & \textbf{94.95 (0.42)} & 73.97 (0.48) & 70.72 (0.28) & \multicolumn{1}{c|}{73.76 (0.24)} & 78.35 \\ 
SiloBN~\cite{SiloBN} & 94.53 (0.85) & 75.95 (0.17) & 70.53 (2.03) & \multicolumn{1}{c|}{78.45 (2.16)} & 79.86 \\ 
\textbf{FedFD-A} & 94.76 (1.14) & \textbf{84.20 (0.31)} & \textbf{79.20 (2.02)} & \multicolumn{1}{c|}{\textbf{82.87 (1.76)}} & \textbf{85.26} \\ 
\bottomrule
\end{tabular}}
\label{30trainingclients_1}
\end{subtable}
\vfill
\vspace{2mm}
\begin{subtable}[h]{0.55\textwidth}
\centering
\subcaption{Performance (\%) on non-iid label distribution ($\alpha=0.5$)}
\vspace{-3mm}
\resizebox{1.0\textwidth}{!}{
\begin{tabular}{c|ccccc}
\toprule                                                      
\multirow{2}{*}{Method} & \multicolumn{5}{c}{PACS} \\ \cline{2-6} 
                        & P & A & C & \multicolumn{1}{c|}{S} & Avg. \\ \midrule
FedProx~\cite{FedProx} & 92.25 (0.25) & 71.75 (3.56) & 74.87 (0.61) & \multicolumn{1}{c|}{69.30 (6.14)} & 77.04 \\ 
SiloBN~\cite{SiloBN} & 92.82 (0.13) & 74.78 (1.55) & 75.50 (0.95) & \multicolumn{1}{c|}{71.71 (2.66)} & 78.70 \\ 
\textbf{FedFD-A} & \textbf{93.62 (0.30)} & \textbf{79.54 (0.83)} & \textbf{82.02 (0.15)} & \multicolumn{1}{c|}{\textbf{77.49 (1.17)}} & \textbf{83.17} \\ 
\bottomrule
\end{tabular}}
\label{30trainingclients_2}
\end{subtable}
\vfill
\vspace{2mm}
\begin{subtable}[h]{0.55\textwidth}
\subcaption{Performance (\%) on non-iid label distribution ($\alpha=0.1$)}
\vspace{-3mm}
\centering
\resizebox{1.0\textwidth}{!}{
\begin{tabular}{c|ccccc}
\toprule                                                      
\multirow{2}{*}{Method} & \multicolumn{5}{c}{PACS} \\ \cline{2-6} 
                        & P & A & C & \multicolumn{1}{c|}{S} & Avg. \\ \midrule
FedProx~\cite{FedProx} & 83.17 (0.42) & 62.65 (4.42) & 66.95 (0.95) & \multicolumn{1}{c|}{55.34 (4.33)} & 67.03 \\ 
SiloBN~\cite{SiloBN} & 84.41 (1.40) & 61.62 (0.48) & 69.05 (3.81) & \multicolumn{1}{c|}{59.85 (4.00)} & 68.73 \\ 
\textbf{FedFD-A} & \textbf{92.13 (0.89)} & \textbf{73.29 (2.90)} & \textbf{73.19 (1.12)} & \multicolumn{1}{c|}{\textbf{71.96 (0.34)}} & \textbf{77.64} \\ 
\bottomrule
\end{tabular}}
\label{30trainingclients_3}
\end{subtable}
\vspace{-2mm}
\end{table}
\begin{table}[t]
\caption{Performance (\%) on the scenario where each client has multiple domains.}
\centering
\vspace{-1mm}
\resizebox{0.55\textwidth}{!}{
\begin{tabular}{c|ccccc}
\toprule                                                      
\multirow{2}{*}{Method} & \multicolumn{5}{c}{PACS} \\ \cline{2-6} 
                        & P & A & C & \multicolumn{1}{c|}{S} & Avg. \\ \midrule
FedProx~\cite{FedProx} & 95.84 (0.13) & 81.13 (2.24) & 75.98 (0.48) & \multicolumn{1}{c|}{74.51 (2.32)} & 81.86 \\ 
SiloBN~\cite{SiloBN} & 95.72 (0.13) & 80.91 (0.14) & 75.96 (0.27) & \multicolumn{1}{c|}{75.82 (0.83)} & 82.10 \\ 
\textbf{FedFD-A} & {\bf 97.82 (0.30)} & {\bf 85.11 (0.00)} & {\bf 78.43 (1.30)} & \multicolumn{1}{c|}{{\bf 82.12 (0.67)}} & {\bf 85.35}\\
\bottomrule
\end{tabular}}
\label{multiple_data_single_client}
\vspace{-4mm}
\end{table}

\subsection{Client with Multi-domain Data}
Irrespective of the number of domains present in a single client, local statistics capture the local domains, while global statistics encompass all the domains across all clients. Thus, our proposed method can be conveniently employed to learn client-invariant representations, even when clients possess diverse domains when training. We conduct experiments with the presence of multiple domains on each client during training. In this scenario, we use three clients, with each client having two domains, such as the first client having P and A domains, the second client having A and C domains, and the third client having C and P domains. After federated learning, we evaluate the global model on the S domain.

When a local client contains data from multiple domains, the local model can learn domain-invariant representations without relying on domain generalization algorithms. Consequently, the regularization technique FedProx~\cite{FedProx} achieves high performance in this scenario—superior to the previous setting where each client had data from a single domain—as demonstrated in Table~\ref{multiple_data_single_client}. However, our method shows similar performance to the single-domain data setting, reported in Table 4 in the main paper. This implies that our approach had already been effectively learning domain-invariant representations and achieving strong generalization with single-domain data. Furthermore, this suggests that our method can be successfully applied in practical situations where clients possess data from multiple domains.

\section{Discussions}
\noindent {\bf Q: What motivates clients to participate in improving the performance of unseen domains?}

\noindent {\bf A:} Clients are motivated to participate in performance improvement for unseen domains to create a more robust model that functions effectively across both in-domain and out-of-domain data, thereby enhancing overall system performance and safety.

In practical situations, clients may face test data that aligns with (in-domain) or diverges from (out-of-domain) the training distribution, with out-of-domain data potentially leading to safety concerns. Our federated feature diversification approach enhances performance across both domains through client-invariant learning. Moreover, feature adaptation leverages the test distribution to further improve results in both in-domain and out-of-domain contexts, as shown in Table~\ref{in_domain_out_domain}. Clients are therefore likely to be motivated to invest in additional training costs to create a more robust model.

\begin{table}[t]
\caption{In-domain and out-of-domain accuracy (\%) on PACS.}
\centering
\resizebox{0.32\textwidth}{!}{
\begin{tabular}{c|cc}
\toprule                                          
Method & in-domain & out-of-domain \\ \midrule
FedAvg~\cite{FedAvg} & 91.43 & 77.35 \\ 
\textbf{FedFD} & 94.53 & 84.07 \\ 
\textbf{FedFD-A} & \textbf{95.20} & \textbf{85.53} \\ 
\bottomrule
\end{tabular}}
\label{in_domain_out_domain}
\end{table}

\begin{table}[t]
\caption{Performance (\%) of SiloBN and FedIG-A with the same training time.}
\centering
\resizebox{0.55\textwidth}{!}{
\begin{tabular}{c|ccccc}
\toprule                                                      
\multirow{2}{*}{Method} & \multicolumn{5}{c}{PACS} \\ \cline{2-6} 
                        & P & A & C & \multicolumn{1}{c|}{S} & Avg. \\ \midrule
SiloBN~\cite{SiloBN} w/ long local iterations & 93.03 & 76.17 & 75.35 & \multicolumn{1}{c|}{78.69} & 80.81 \\ 
SiloBN~\cite{SiloBN} w/ long total rounds & 93.09 & 76.76 & 76.65 & \multicolumn{1}{c|}{75.43} & 81.23 \\ 
\textbf{FedFD-A} & \textbf{94.24} & \textbf{84.30} & \textbf{79.80} & \multicolumn{1}{c|}{\textbf{83.79}} & \textbf{85.53}\\
\bottomrule
\end{tabular}}
\label{same_training_time}
\end{table}

\noindent {\bf Q: How does FedFD-A compare to SiloBN in terms of performance when training time is the same?}

\noindent {\bf A: }
FedFD-A outperforms SiloBN, even with SiloBN's increased training time, due to the superior client-invariant representation learning of FedFD-A.

FedFD-A requires a 50\% increase in training time compared to SiloBN.
For a fair comparison, we conduct additional experiment on SiloBN with more iterations, reported in Table~\ref{same_training_time}.
Experiments show that SiloBN, even when given three times more iterations (600) or rounds (120) than FedFD-A, achieves only marginal performance improvements in its extended setting. Despite this, FedFD-A still outperforms SiloBN by a considerable margin, emphasizing that local models in SiloBN cannot fully capture client-invariant representations amid domain shifts among clients.

\begin{table}[t]
\caption{Performance (\%) on FL models with short local iterations.}
\centering
\resizebox{0.55\textwidth}{!}{
\begin{tabular}{c|ccccc}
\toprule                                                      
\multirow{2}{*}{Method} & \multicolumn{5}{c}{PACS} \\ \cline{2-6} 
                        & P & A & C & \multicolumn{1}{c|}{S} & Avg. \\ \midrule
FedProx~\cite{FedProx} & 92.40 (0.47)&	74.93 (0.03)&	72.25 (0.03)	&\multicolumn{1}{c|}{72.28 (0.22)}&	77.88 \\ 
SiloBN~\cite{SiloBN} & 93.09 (0.34)&	75.42 (0.17)	&73.48 (0.61)&	\multicolumn{1}{c|}{76.97 (2.21)}	&79.74 \\ 
\textbf{FedFD-A} & \textbf{93.56 (0.47)} &	\textbf{84.89 (0.24)} &	\textbf{79.91 (0.60)} &	\multicolumn{1}{c|}{\textbf{83.98 (0.49)}}&	\textbf{85.58} \\
\bottomrule
\end{tabular}}
\label{short_local_iteration}
\end{table}

\noindent {\bf Q: Can local models avoid overfitting to their specific domains if the models are aggregated frequently?}

\noindent {\bf A: }
Even with frequent aggregation and shorter local epochs, local models may still overfit to their domains.

We have conducted experiments with local iterations that are ten times shorter and total rounds that are ten times longer than usual. The results suggest that frequent aggregation does not necessarily prevent overfitting. As evidenced in Table~\ref{short_local_iteration}, overfitting in local models to their respective domains still occurs. Therefore, to prevent overfitting to individual domains, relying solely on frequent model aggregation appears to be insufficient. Our proposed methods, however, have demonstrated robustness, maintaining good performance regardless of the local iteration length.

\noindent {\bf Q: Can our method not be used in a base architecture that does not contain batch normalization layers?}

\noindent {\bf A: }
While most CNN-based architectures incorporate batch normalization layers to enhance model generalization, our method can still be applied to architectures without these layers by integrating batch normalization layers into them. In the DomainNet experiments, we inserted batch normalization layers into AlexNet and conducted tests, a practice also employed in other papers. Through these experiments, we demonstrated that by inserting batch normalization layers into the existing architecture, our method can be effectively implemented.

\section{Pseudo-code for reproducibility}
The pseudo-codes for federated feature diversification and feature adaptaion are presented in Table~\ref{pseudo_1} and~\ref{pseudo_2}.

\begin{table*}[t]
\centering
\caption{Pseudo-code for FedFD-A training. The equation numbers are all from the main paper.}
\resizebox{0.67\linewidth}{!}{
\begin{tabular}{l}
\toprule
 Global weights $\theta_{G}$, $\phi_{G}$, $\varphi_{G}$, Local clients' weights $\{\theta_{k}, \phi_{k}, \varphi_{k}\}_{k=1}^{K}$, \\
 Total round $T_{max}$, Total local iteration $E_{max}$; \\ \midrule
 {\bf Server executes:} \\
 \hspace{10mm}initialize $\theta_{G}$, $\phi_{G}$, $\varphi_{G}$; \\
 \hspace{10mm}{\bf for} each round $t = 1, ..., T_{max}$ {\bf do} \\
 \hspace{20mm}$S_{t}$ $\leftarrow$ (random set of m clients);\\
 \hspace{20mm}{\bf for} each client $k \in S_{t}$ {\bf in parallel do} \\
 \hspace{30mm}{\bf Load} $\theta_{k}$, ${\phi_{k}}$, $\varphi_{k}$ $\leftarrow$ LocalUpdate($k$, $\theta_{G}$, ${\phi_{G}}$, $\varphi_{G}$);\\
\hspace{20mm}{\bf Update} global weights $\theta_{G}$, ${\phi_{G}}$, $\varphi_{G}$ by FedAvg~\cite{FedAvg};\\
\hspace{10mm}{\bf Output}: $\theta_{G}$, ${\phi_{G}}$, $\varphi_{G}$.\\
 \\
 {\bf function} LocalUpdate($k$, $\theta_{G}$, ${\phi_{G}}$, ${\varphi_{G}}$): // Run on client k\\
 \hspace{10mm}{\bf Load} $\theta_{k}$, ${\phi_{k}}$, $\varphi_{k}$ $\leftarrow$ $\theta_{G}$, ${\phi_{G}}$, $\varphi_{G}$;\\
 \hspace{10mm}{\bf for} each local iteration $i = 1, ..., E_{max}$ {\bf do} \\
 \hspace{20mm}{\bf Shuffle} training set $D_{k}$; \\
 \hspace{20mm}{\bf Fetch} mini batch $\{x_{i, k}, y_{i, k}\}_{i=1}^{n}$ from $D_{k}$;\\ \\
 \hspace{20mm}// Train the main network \\
 \hspace{20mm}{\bf Obtain} loss $\mathcal{L}_{CE}$ by Eq. 1;\\
 \hspace{20mm}{\bf Obtain} augmented features $\{f_{i, \Delta}\}_{i=1}^{n}$ with $\theta_{G}$ by Eq. 2;\\
 \hspace{20mm}{\bf Obtain} loss $\mathcal{L}_{CAFL}$, $\mathcal{L}_{CACL}$ by Eq. 3 and 4;\\
 \hspace{20mm}{\bf Update} local weights $\theta_{k}$, ${\phi_{k}}$ by minimizing Eq. 5;\\ \\
 \hspace{20mm}// Train the instance feature adapter \\
 \hspace{20mm}{\bf Obtain} features with estimated statistics by Eq. 6 and 7;\\
 \hspace{20mm}{\bf Update} local weights $\varphi_{k}$ by minimizing Eq. 1;\\
 \hspace{10mm}{\bf Output}: $\theta_{k}$, ${\phi_{k}}$, $\varphi_{k}$.\\
 \bottomrule
\end{tabular}
\label{pseudo_1}}
\end{table*}

\begin{table*}[t]
\centering
\caption{Pseudo-code for FedFD-A inference. The equation numbers are all from the main paper.}
\resizebox{0.55\linewidth}{!}{
\begin{tabular}{l}
\toprule
 Global weights $\theta_{G}$, ${\phi_{G}}$, $\varphi_{G}$, Test client's weights $\theta_{t}$, ${\phi_{t}}$, $\varphi_{t}$;\\ \midrule
 {\bf Deploy to test client:} \\
 \hspace{10mm}{\bf Load} $\theta_{t}$, ${\phi_{t}}$, $\varphi_{t}$ $\leftarrow$ $\theta_{G}$, ${\phi_{G}}$, $\varphi_{G}$;\\
 \hspace{10mm}{\bf Obtain} test set $D_{t}$; \\
 \hspace{10mm}{\bf for} each test forward $i = 1, ..., n_{t}$ {\bf do} \\
 \hspace{20mm}{\bf Fetch} $x_{i, t}$ from $D_{t}$; \\
 \hspace{20mm}{\bf Obtain} $f_{i, t}$ by Eq. 6; \\
 \hspace{20mm}{\bf Obtain} the prediction with ${\phi_{t}}$; \\
\bottomrule
\end{tabular}
\label{pseudo_2}}
\end{table*}

\clearpage  

\end{document}